\documentclass{article}
\usepackage{iclr2027_conference,times}

\usepackage{amsmath,amsfonts,bm}

\def\eqref#1{equation~\ref{#1}}

\def\1{\bm{1}}

\DeclareMathAlphabet{\mathsfit}{\encodingdefault}{\sfdefault}{m}{sl}
\SetMathAlphabet{\mathsfit}{bold}{\encodingdefault}{\sfdefault}{bx}{n}

\usepackage{latexsym}
\usepackage[T1]{fontenc}
\usepackage[utf8]{inputenc}
\usepackage{microtype}
\usepackage{inconsolata}
\usepackage{graphicx}
\usepackage{hyperref}
\usepackage{url}

\usepackage{booktabs}
\usepackage{multirow}
\usepackage{amsmath,amssymb}
\usepackage[most]{tcolorbox}
\usepackage{wrapfig}
\usepackage{pgfplots}
\usepgfplotslibrary{groupplots,fillbetween}
\pgfplotsset{compat=1.18}
\definecolor{CorrectGreen}{RGB}{28,165,122}
\definecolor{IncorrectOrange}{RGB}{214,91,0}
\definecolor{IncompletePurple}{RGB}{116,108,184}
\definecolor{BaseBlue}{RGB}{31,119,179}
\definecolor{RegularOrange}{RGB}{245,133,24}
\definecolor{ThinkingGreen}{RGB}{130,194,101}
\definecolor{InstructRed}{RGB}{225,87,89}
\definecolor{AdjGold}{RGB}{230,184,56}

\newtcolorbox{promptbox}[1][]{
  enhanced,
  breakable,
  colback=gray!3,
  colframe=gray!55,
  boxrule=0.6pt,
  arc=4pt,
  left=8pt,right=8pt,top=7pt,bottom=7pt,
  title={Prompt Template},
  #1,
}

\newcommand{\psection}[1]{\textbf{#1}\par}

\title{LLMs Think with DAGs: Probing the Latent Graph Structure of Reasoning}

\author{Tianjun Zhong\textsuperscript{1,2}, Linyang He\textsuperscript{1}, Ziyang Li\textsuperscript{2}, Nima Mesgarani\textsuperscript{1}\\ \textsuperscript{1}Columbia University \quad \textsuperscript{2}Johns Hopkins University\\ \texttt{tzhong4@jhu.edu, linyang.he@columbia.edu}\\ \texttt{ziyang@cs.jhu.edu, nima@ee.columbia.edu} }

\iclrfinalcopy 
\begin{document}
\maketitle
\fancyhead{}
\renewcommand{\headrulewidth}{0pt}

\begin{abstract}

Large language models often externalize reasoning as linear chains, 
but many reasoning problems can be more naturally modeled as directed acyclic graphs (DAGs), 
where intermediate conclusions branch, merge, and are reused. 
Whether such graph structure is reflected in model internals remains unclear.
We introduce \emph{Reasoning DAG Probing} to answer this question.
We formalize each premise, intermediate conclusion, and final answer as a DAG node, 
and train lightweight probes to predict node depth, pairwise distance, and adjacency from LLM hidden states.
Across logical, mathematical, and code reasoning benchmarks, 
we find that DAG structure is meaningfully encoded in LLM representations: 
it peaks in intermediate layers, 
with later layers preferentially recover deeper nodes and longer-range reasoning dependencies, 
and becomes stronger with model scale.
DAG structure also progressively emerges during autoregressive generation 
in ways that anticipate answer correctness, 
while steering activations toward the ground-truth DAG structure can convert some originally incorrect generations into correct ones.
These results suggest that LLM reasoning is not purely sequential, 
but reflects measurable latent graph structure.
\end{abstract}
\begin{figure*}[t]
\centering
\includegraphics[width=\textwidth]{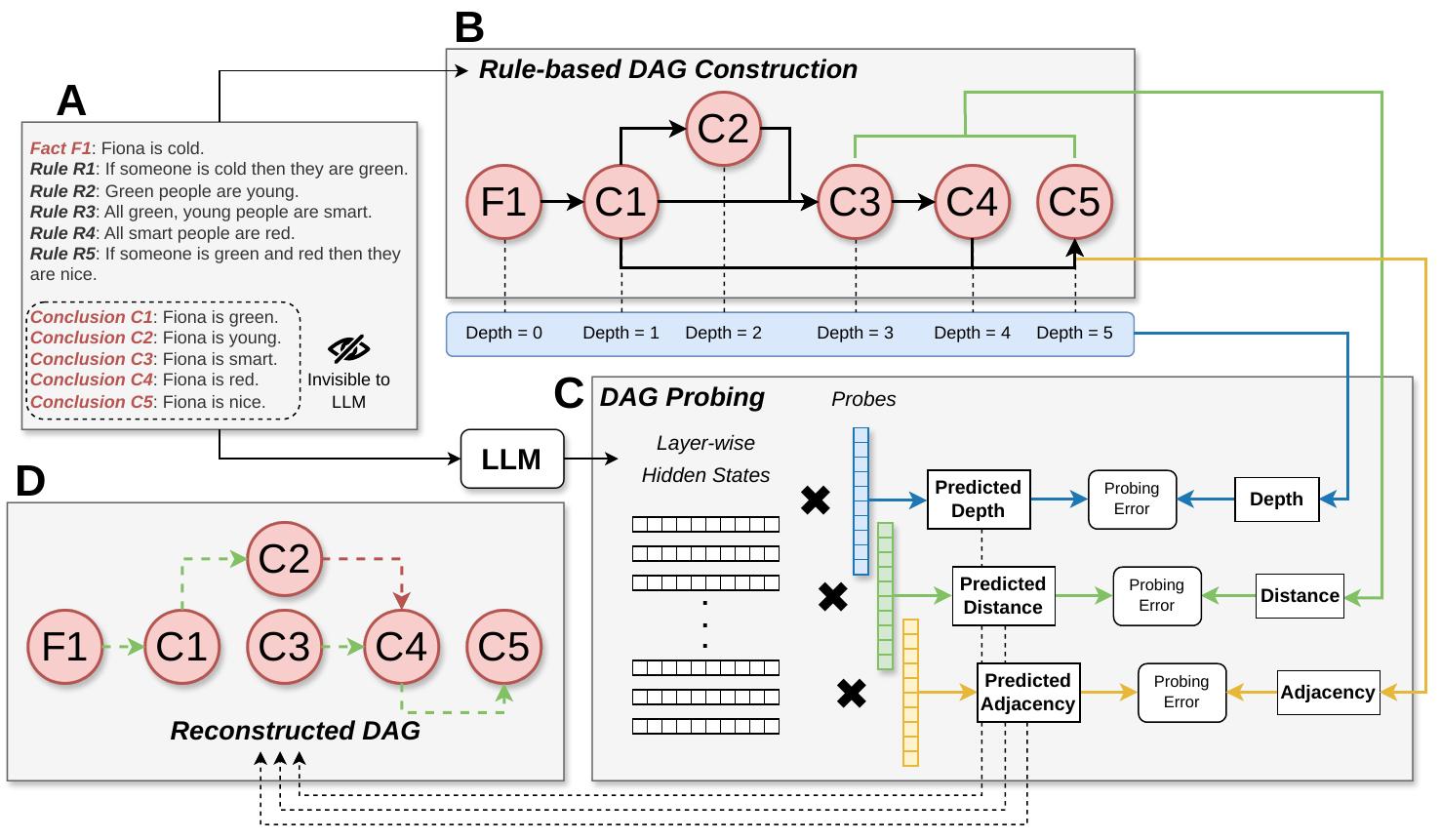}
\caption{Overall pipeline of Reasoning DAG Probing.
\textbf{(A) Reasoning Data:} The input consists of a multi-step reasoning problem expressed in natural language (e.g., facts, rules, and conclusions from ProofWriter).
\textbf{(B) Data Construction:} The multi-step reasoning problem is formalized as a Directed Acyclic Graph (DAG), where nodes represent premises, intermediate conclusions, or final conclusion, and edges denote dependency relations.
\textbf{(C) Structural Probing:} The textual realization of the reasoning problem is processed by the LLM.
Hidden states corresponding to each node are extracted.
Linear probes are trained on these representations to predict each node's hierarchical depth, pairwise node distance, and adjacency.
\textbf{(D) Graph Reconstruction:} The reasoning structure is recovered by inferring edges based on the predicted depth, distance, and adjacency.}
\label{fig:pipeline}
\end{figure*}

\section{Introduction}
Large language models (LLMs) can solve many multi-step reasoning tasks, especially when prompted to externalize intermediate steps with chain-of-thought (CoT) or related rationales \citep{wei2022chain}.
However, the relationship between these textual traces and the underlying computation is unresolved.
Generated explanations may be incomplete, post hoc, or strategically produced, and therefore cannot by themselves serve as a faithful description of the model's internal algorithm \citep{turpin2023language,lanham2023measuring,barez2025chain}.
This motivates methods that localize and quantify structure directly inside model activations.

A key limitation of the dominant CoT framing is that it linearizes reasoning.
Real reasoning often has graph structure \citep{yao-etal-2024-got}.
Multiple premises can jointly support an intermediate conclusion \citep{dalvi-etal-2021-explaining}, intermediate results can be reused, and long derivations can branch and later merge.
These properties are naturally captured by directed acyclic graphs (DAGs), whereas a single chain can collapse dependencies and obscure compositional structure.

This paper asks whether LLM's internal representations reflect this graph view.
Rather than probing hidden states for isolated variables, we probe for the geometry and dependency structure of a reasoning DAG.
Building on the structural probe methodology of \citet{hewitt-manning-2019-structural}, we train low-capacity probes on frozen hidden states to recover three graph-theoretic properties that summarize DAG structure:
(i) node depth, which captures each reasoning state's relative position in the reasoning hierarchy,
(ii) pairwise node distance, which measures how far two states lie from each other in the dependency structure, and
(iii) pairwise node adjacency, which indicates whether two states are directly connected by an edge in the reasoning graph.
If these properties can be accurately recovered by a simple probe at a given layer, then the corresponding layer contains a linearly accessible encoding of reasoning DAG structure.

We use this lens to address four questions.
First, do LLM hidden states encode a structured reasoning DAG at all, beyond what is present in the explicit rationale text?
Second, where and how does this structure emerge across depth in an LLM?
Third, how does DAG recoverability vary as LLMs generate their answers, and across training recipe and model scale?
Fourth, is the encoded DAG structure causally involved in LLM reasoning, or is it merely passively decodable from hidden states?

In summary, our contributions are fourfold:
\begin{enumerate}
\setlength{\itemsep}{0pt}
\setlength{\parsep}{0pt}
\setlength{\topsep}{0.25em}
\item We formalize multi-step reasoning in logical deduction, arithmetic word problems, and code-generation tasks as directed acyclic graphs (DAGs), and define probing targets based on node depth, pairwise distance, and direct dependency between nodes.
\item We propose \emph{Reasoning DAG Probing}, a framework that adapts structural probes to recover DAG properties from LLM hidden states and uses them to reconstruct reasoning DAGs with layerwise resolution.
\item We empirically characterize when and where reasoning DAG structure becomes recoverable, analyzing its dependence on node depth, edge span, model scale, and training recipe.
\item We study latent DAG structure during autoregressive generation, showing that recoverability tracks answer formation. Generation-time interventions further provide evidence that the probed structure contributes to reasoning behavior.

\end{enumerate}
\section{Methods}

\subsection{Dataset and DAG Construction}
We evaluate Reasoning DAG Probing across three reasoning settings: logical deduction, arithmetic word problems, and algorithmic code generation.
We use ProofWriter \citep{tafjord2021proofwriter} as the primary benchmark for demonstrating the framework because its rule-based natural language inference examples provide explicit, interpretable proof structures.
Each ProofWriter example (see Appendix~\ref{sec:proofwriter_example}) consists of a theory (true statements and rules),
a query (a statement whose truth value is inferred from the theory),
and a proof that derives the answer through rule applications.
From each proof, we construct a DAG whose nodes are proof statements, including premises, intermediate conclusions, and the final answer, with directed edges from the premise nodes of each rule application to its conclusion node.
We use each node's statement text as its textual realization for LLM encoding.

We also evaluate on GSM8K \citep{cobbe2021gsm8k}, a mathematical word-problem benchmark, and TACO \citep{li2023taco}, an algorithmic code-generation benchmark.
For these settings, we construct task-specific reasoning DAGs from worked solutions and apply the same probing framework.
Dataset details are provided in Appendices~\ref{sec:gsm8k} and~\ref{sec:taco}.

\subsection{Reasoning Graph Properties}
\paragraph{Depth.}
Let $G=(V,E)$ be a directed acyclic graph, where $V$ is the set of nodes and $E$ is the set of directed edges, with a designated sink node $s$ corresponding to the final answer.
For any node $v \in V$, let
\[
d_{\text{raw}}(v, s) = \max_{p \in \mathcal{P}(v \rightarrow s)} |\;p\;|
\]
denote the length of the longest directed path from $v$ to the sink $s$, where $\mathcal{P}(v \rightarrow s)$ is the set of all directed paths from $v$ to $s$.
We use the longest path, not the shortest, to reflect the maximum number of reasoning steps involving a node, and to avoid skipping premises in settings where multiple premises must jointly contribute to a conclusion (see Figure~\ref{fig:dag_example} and Appendix~\ref{paragraph:why_longest}).
Since $d_{\text{raw}}(s,s)=0$ and larger values correspond to more peripheral premises, we define a normalized depth measure
\[
\operatorname{depth}(v) = \max_{u \in V} d_{\text{raw}}(u,s) - d_{\text{raw}}(v,s),
\]
so that the sink $s$ has maximal depth and premise nodes are shallower.

\paragraph{Distance.}
For two distinct nodes $u,v \in V$, we define the pairwise symmetric distance by dropping edge direction after determining which node reaches the other:
\[
\operatorname{dist}(u,v) =
\begin{cases}
d_{\text{raw}}(u,v), & \text{if } u \leadsto v, \\
d_{\text{raw}}(v,u), & \text{if } v \leadsto u, \\
\text{undefined}, & \text{otherwise}.
\end{cases}
\]
Here $u \leadsto v$ denotes the existence of a directed path from $u$ to $v$.
Because $G$ is acyclic, at most one of the first two cases can hold for $u \ne v$.
Pairs in different branches with no directed path in either direction do not have a gold distance label, so they are excluded when training and evaluating the distance probe as a standalone probing task.

\paragraph{Adjacency.}
For two nodes $u,v \in V$, we define the pairwise adjacency
\[
\operatorname{adj}(u,v) =
\begin{cases}
1, & \text{if } (u,v) \in E \text{ or } (v,u) \in E, \\
0, & \text{otherwise}.
\end{cases}
\]
That is, pairwise adjacency indicates whether two nodes are directly connected by an edge in the reasoning graph, independent of edge direction.

Together, node depth, pairwise distance, and pairwise adjacency capture node-level, relational, and local connectivity structure in the DAG, which we probe from LLM hidden states.

\subsection{Probing Setup}
\label{subsec:probing_setup}
\paragraph{Representation.}
For each reasoning DAG node $v$, we construct a textual input by concatenating the full theory $\mathcal{T}$ and the node's textual realization $x_v$.
The model sees only the problem context and the single target statement.
Thus, probing tests whether hidden states recover DAG structure without being given an explicit step-by-step solution.
Given a pretrained language model with frozen parameters, this input is tokenized with offset mappings to identify the token span corresponding to $x_v$.
For each probed layer $\ell$, we extract the hidden states $\{h^{(\ell)}_t : t \in \mathrm{span}(v)\}$ and mean-pool them to obtain a single node representation
\[
\mathbf{z}^{(\ell)}_v = \frac{1}{|\mathrm{span}(v)|} \sum_{t \in \mathrm{span}(v)} h^{(\ell)}_t \in \mathbb{R}^d.
\]
During generation-time analysis, we additionally evaluate a native-span variant, 
where probes read from the model's own generated mentions of DAG nodes instead of appended node text.

\paragraph{Probes.}
All three probes are parameterized as low-rank linear maps with rank $k=1$ and no bias, and are trained with AdamW on frozen representations.
Model selection is performed based on development loss with task-specific hyperparameters.

The depth probe assigns a scalar prediction $\hat{\operatorname{depth}}(v) = \mathbf{w}^\top \mathbf{z}_v$ to each node and is trained using a pairwise ranking objective: for nodes $u,v$ in the same graph with $\operatorname{depth}(u) = \operatorname{depth}(v)+1$, we minimize the following to assign higher values to deeper nodes:
\[
\mathcal{L}_{\text{depth}} = \mathrm{softplus}\bigl( -(\hat{\operatorname{depth}}(u) - \hat{\operatorname{depth}}(v)) \bigr).
\]

The distance probe operates on comparable pairs of nodes and predicts graph distance from the absolute difference of their representations, $\lvert \mathbf{z}_u - \mathbf{z}_v \rvert$, using a linear map and mean-squared error loss against the annotated pairwise distances.

The adjacency probe likewise operates on pairs of nodes.
It computes a scalar score
\[
s_{uv} = \mathbf{w}^\top \lvert \mathbf{z}_u - \mathbf{z}_v \rvert,
\]
which is passed through a sigmoid nonlinearity to obtain the predicted probability of adjacency $\hat{\operatorname{adj}}(u,v)$, and is trained with binary cross-entropy loss against the gold adjacency label.
To account for class imbalance, we use a positive-class weighting term in the binary cross-entropy loss.

\paragraph{Baselines.}
We consider three baseline variants to isolate the contribution of contextual and structural information: (i) a node-only setting where $\mathcal{T}$ is omitted and representations are extracted from $x_v$ alone; (ii) a bag-of-words baseline that replaces contextual representations with fixed-dimensional lexical features; and (iii) a label-shuffled control in which gold depth, distance, and adjacency annotations are randomly permuted.

\subsection{DAG Reconstruction}
\label{subsec:reconstruct_algo}
Given probe predictions for node depth, pairwise distance, and pairwise adjacency, we reconstruct an approximate reasoning DAG using a threshold-based procedure.
For each graph, we first identify the predicted sink node
\[
\hat{s} = \arg\max_{v \in V} \hat{\operatorname{depth}}(v).
\]

We then consider all unordered node pairs $\{u,v\}$ with predicted distance $\hat{\operatorname{dist}}(u,v)$ and adjacency probability $\hat{\operatorname{adj}}(u,v)$.
Each pair is oriented according to the predicted depth ordering, with edges directed from shallower to deeper node.
Since edge orientation is induced by the predicted depth ordering, the resulting graph is acyclic by construction.

Decoding proceeds in two steps.
First, for each non-sink node, we add its best outgoing candidate edge, chosen by highest predicted adjacency probability with lower predicted distance used as a tie-breaker.
Second, we add any remaining candidate edge satisfying
\[
\hat{\operatorname{dist}}(u,v) \le \tau_{\text{dist}}
\quad \text{and} \quad
\hat{\operatorname{adj}}(u,v) \ge \tau_{\text{adj}}.
\]
The thresholds $\tau_{\text{dist}}$ and $\tau_{\text{adj}}$ are tuned on the development set to maximize reconstruction performance.

\subsection{Model Generation}
Our primary experiments use the Qwen3 family of LLMs \citep{yang2025qwen3}.
For each example, we prompt the model with the task input and ask it to produce a reasoning trace followed by an explicit final answer in a task-specific format.
Generation is performed with stochastic decoding enabled and a preset token budget.
We parse the final answer and record correctness for generation-time analyses.

\subsection{DAG Steering Setup}
\label{sec:steering_method}
To test whether recoverable DAG structure influences the model's subsequent generation, 
we perform a generation-time activation intervention on Qwen3-4B (details in Appendix \ref{app:dag_steering_details}).
For a cutoff \(c\), defined as the first \(c\) tokens of an generated trace, 
we align the partial trace to the gold DAG, 
select native token spans from the partial trace where DAG nodes are mentioned, 
and optimize an additive perturbation to the corresponding hidden states of the selected spans.
The perturbation is optimized to reduce a differentiable probe loss over node depth, pairwise distance, and adjacency, using weights \(1{:}1{:}1\).
Each intervention uses a single gradient update, projected to a perturbation norm cap expressed as a fraction of the selected hidden-state norm, and generation then continues from the intervened prefix.
For multi-cutoff steering, we repeat this procedure at later cutoffs, recomputing the native span alignments before each intervention.
We tune the norm caps and compare a single intervention at \(c=384\) with repeated interventions at \(c=384,1024\) and \(c=384,1024,2048\).
\begin{figure*}[t]
\centering
\input{fig/probe_reconstruct_main}
\vspace{-1.5em}
\caption{Reasoning DAG probing and reconstruction results for Qwen3-14B on ProofWriter, computed before any generated tokens are appended.
\textbf{(a)} Layerwise probing performance for node depth, pairwise node distance, and pairwise adjacency.
\textbf{(b)} Peak probe performance across layers for the main method and three baselines.
\textbf{(c)} DAG reconstruction F1 at layer 25 across distance and adjacency thresholds, showing stable reconstruction quality over a broad range of decoding settings.}
\label{fig:probe_reconstruct}
\end{figure*}

\section{Results}
\subsection{Layerwise Emergence of DAG Geometry}
We evaluate three probing metrics across layers.
\emph{Depth Spearman} measures the mean per-graph Spearman correlation between predicted and gold node depths, capturing whether nodes are correctly ordered in the reasoning DAG.
\emph{Dist Spearman} measures the same for pairwise node distances, assessing recovery of relative graph geometry.
\emph{Adjacency F1} measures binary prediction of pairwise adjacency after thresholding, assessing recovery of direct graph edges.

As shown in Figure~\ref{fig:probe_reconstruct}(a), all three metrics exhibit a non-uniform layerwise pattern in Qwen3-14B.
Performance improves rapidly in the early layers, reflecting the progressive incorporation of contextual information beyond surface form, while the earliest layers remain weak and dominated by lexical features \citep{voita-etal-2019-analyzing, he2025far}.
A broad band of intermediate layers achieves the strongest recovery of DAG geometry, with all three metrics peaking concurrently around layer 25.
In the final layers, performance mildly declines, possibly reflecting a shift toward objectives less aligned with reasoning structure.

\subsection{Failure of Baselines to Encode Reasoning DAG Geometry}
To assess whether our probes recover genuine reasoning structure rather than superficial correlations, we compare against three baseline conditions that selectively remove or destroy DAG information: a \emph{node-only} setting without access to the surrounding theory, a \emph{bag-of-words} baseline that replaces contextual embeddings with lexical features, and a \emph{label-shuffled} control that breaks the alignment between representations and graph structure.

Figure~\ref{fig:probe_reconstruct}(b) summarizes the peak probing performance across layers for each condition.
Across depth, distance, and adjacency metrics, the main method substantially outperforms all baselines.
Node-only retains limited residual signal, suggesting that isolated node text can weakly correlate with graph structure but cannot recover full DAG geometry.
Bag-of-words degrades further, indicating that lexical statistics alone do not encode reasoning structure.
Label shuffling collapses performance nearly entirely, confirming that probe success depends on alignment between hidden representations and the gold DAG instead of probe expressivity alone.
The layerwise baseline results in Figure~\ref{fig:layerwise_baselines} are also largely flat, with no clear upward trend, indicating the absence of progressive reasoning structure.
Together, these controls show that recoverable DAG structure arises from contextual representations over the full theory, not shallow textual features or label artifacts.

\subsection{DAG Reconstruction Performance}
Following Section~\ref{subsec:reconstruct_algo}, we reconstruct reasoning DAGs and evaluate them by comparing predicted edges $\hat{E}$ against gold edges $E$.
For each graph, we compute edge-level precision, recall, and DAG-F1,
\[
\makebox[\linewidth][c]{$
\begin{gathered}
\mathrm{Precision} = \frac{|\hat{E}\cap E|}{|\hat{E}|},
\qquad
\mathrm{Recall} = \frac{|\hat{E}\cap E|}{|E|}, \\
\text{DAG-F1} = \frac{2 \cdot \mathrm{Precision}\ \cdot \mathrm{Recall}}{\mathrm{Precision}+\mathrm{Recall}}.
\end{gathered}
$}
\]
and average these metrics across the evaluation set.

Figure~\ref{fig:probe_reconstruct}(c) shows peak DAG-F1 across layers as a function of the distance threshold $\tau_{\text{dist}}$ and adjacency threshold $\tau_{\text{adj}}$.
Reconstruction remains stable across a broad threshold region, with the highest DAG-F1 at moderate adjacency thresholds and moderate-to-large distance thresholds.
This suggests that reconstruction benefits from admitting longer-range dependency candidates while filtering low-confidence edges, consistent with ProofWriter graphs where valid reasoning edges can span several inference steps (Figure~\ref{fig:dag_example}).

To illustrate how reconstruction quality evolves across layers, Figure~\ref{fig:case_study} presents a visual case study.
Early layers show unstable depth ordering and noisy connectivity; intermediate layers best recover node ordering and gold dependencies; and later layers preserve coarse structure while losing edge-level precision.
This mirrors the intermediate-layer peak and mild late-layer decline in Figure~\ref{fig:probe_reconstruct}(a).
\begin{figure*}[t]
\centering
\begin{tikzpicture}
\begin{axis}[
  width=0.9\textwidth,
  height=3.5cm,
  xmin=-0.5, xmax=40.5,
  ymin=-0.5, ymax=5.5,
  xtick={0,5,10,15,20,25,30,35,40},
  xticklabels=\empty,
  ytick={0,1,2,3,4,5},
  xlabel={},
  ylabel={Gold depth},
  title={{\small\textbf{(a)} Depth MAE by node depth}},
  title style={font=\small, yshift=-0.3cm},
  tick align=outside,
  xticklabel style={font=\scriptsize},
  yticklabel style={font=\scriptsize},
  label style={font=\scriptsize},
  axis line style={black!70},
  colormap={BluesR}{
    rgb255=(8,54,120)
    rgb255=(8,74,145)
    rgb255=(23,100,171)
    rgb255=(46,126,188)
    rgb255=(74,152,201)
    rgb255=(108,174,214)
    rgb255=(148,196,223)
    rgb255=(183,212,234)
    rgb255=(208,225,242)
    rgb255=(227,238,249)
    rgb255=(247,251,255)
  },
  colorbar,
  colorbar style={
    ylabel={MAE},
    ylabel style={font=\scriptsize, yshift=0.1cm},
    yticklabel style={font=\scriptsize},
    width=0.12cm,
    xshift=-0.12cm
  },
  point meta min=0,
  point meta max=10,
]
\addplot[
  matrix plot*,
  draw=white,
  line width=0.08pt,
  mesh/cols=41,
  point meta=explicit,
] table[x=layer,y=depth_bin,meta=mae,col sep=comma]{data/depth_binned_mae_qwen14b.csv};
\end{axis}

\begin{axis}[
  at={(0,-0.47cm)},
  anchor=north west,
  width=0.9\textwidth,
  height=3.5cm,
  xmin=-0.5, xmax=40.5,
  ymin=0.5, ymax=5.5,
  xtick={0,5,10,15,20,25,30,35,40},
  ytick={1,2,3,4,5},
  xlabel={LLM Layer},
  xlabel style={font=\scriptsize, yshift=0.2cm},
  ylabel={Edge span},
  title={{\small\textbf{(b)} Adjacency AP by edge span}},
  title style={font=\small, yshift=-0.3cm},
  tick align=outside,
  xticklabel style={font=\scriptsize, yshift=5pt},
  yticklabel style={font=\scriptsize},
  label style={font=\scriptsize},
  axis line style={black!70},
  colormap={BluesR}{
    rgb255=(8,54,120)
    rgb255=(8,74,145)
    rgb255=(23,100,171)
    rgb255=(46,126,188)
    rgb255=(74,152,201)
    rgb255=(108,174,214)
    rgb255=(148,196,223)
    rgb255=(183,212,234)
    rgb255=(208,225,242)
    rgb255=(227,238,249)
    rgb255=(247,251,255)
  },
  colorbar,
  colorbar style={
    ylabel={Normalized AP},
    ylabel style={font=\scriptsize, yshift=0.1cm},
    yticklabel style={font=\scriptsize},
    width=0.12cm,
    xshift=-0.12cm
  },
  point meta min=0,
  point meta max=1,
]
\addplot[
  matrix plot*,
  draw=white,
  line width=0.08pt,
  mesh/cols=41,
  point meta=explicit,
] table[x=layer,y=depth_gap,meta=ap_minmax,col sep=comma]{data/edge_depth_gap_ap_minmax_qwen14b.csv};
\end{axis}
\end{tikzpicture}
\caption{\textbf{(a)} Depth-probe mean absolute error (MAE) grouped by node depth across layers.
Rows correspond to gold graph depth bins (0 = shallowest, 5 = deepest), and columns correspond to model layers.
Lower values (darker) indicate more accurate depth prediction.
\textbf{(b)} Adjacency-probe average precision grouped by edge span (depth gap) across layers.
Rows correspond to depth-gap bins (1 = smallest, 5 = largest).
Higher values (lighter) indicate more accurate adjacency prediction.}
\label{fig:depth_binned_mae}
\end{figure*}

\subsection{Recoverability as a Function of Node Depth and Edge Span}
To examine depth-dependent recoverability, we group nodes by gold graph depth (0–5) and evaluate depth probe performance within each group.
Since nodes in a bin share the same gold depth, rank-based metrics such as Spearman correlation are undefined, so we report mean absolute error (MAE) between predicted and gold depths. 
Figure~\ref{fig:depth_binned_mae}(a) reveals a clear interaction between layer and node depth: optimal recoverability progresses from shallow nodes in earlier layers to deeper nodes in later layers.
This produces a dark diagonal banding pattern in the heatmap, indicating that deeper layers preferentially improve representations of deeper reasoning steps, while mild late-layer degradation for shallow nodes suggests a shift in representational emphasis rather than uniform gains.

To examine span-dependent adjacency recoverability, we group node pairs by their edge span (depth gap) and evaluate adjacency probe within each group (details in Appendix~\ref{app:adj_norm}).
Figure~\ref{fig:depth_binned_mae}(b) shows a roughly diagonal transition into the high-value region: short-span edges become recoverable earlier, whereas edges spanning larger depth gaps become recoverable only in deeper layers.
Compared with the node-depth pattern in Figure~\ref{fig:depth_binned_mae}(a), this later transition suggests that pairwise relational structure emerges after local node position.
Together, these results are consistent with a progressive reallocation of representational capacity toward increasingly global dependencies.

\subsection{DAG Recoverability and Generation Correctness}
\label{sec:gen_correct}
We next ask whether DAG recoverability tracks the model's own autoregressive generation dynamics.
We use two complementary views (details in Appendix~\ref{app:dag_gen_details}):
appended-node probing appends node text after partial generations at different token cutoffs, mirroring the setup in Section~\ref{subsec:probing_setup};
native-span probing instead reads activations at tokens where the model itself mentions the corresponding nodes.

Figure~\ref{fig:temporal_generation}(a) shows Qwen3-14B on ProofWriter, grouped by final answer correctness.
Correct and incorrect generations begin with similar DAG-F1, but diverge during decoding: correct generations show a sustained increase in recovered DAG structure, whereas incorrect generations remain relatively flat.
This separation is useful because generated reasoning traces can lead to wrong answers; low or stagnant DAG-F1 helps identify such failures despite the presence of CoT.
Thus, successful answers are associated with progressively clearer internal organization of the underlying reasoning DAG.

Figure~\ref{fig:temporal_generation}(b) compares appended-node DAG-F1, native-span DAG-F1, and correctness at each generated-token cutoff across Qwen3 4B variants.
Both DAG-F1 views anticipate the broad shape of later correctness curves while reflecting the generation style of each variant.
The base model answers early with little extended reasoning, and its DAG-F1 remains nearly flat, suggesting that generation adds little new hierarchical structure.
The standard and instruction-tuned variants reach correctness earlier without long deliberation, whereas the thinking model delays its answer while producing extended CoT.
Across these variants, DAG-F1 reflects the relative timing of correctness gains and the final ordering among models.
For the thinking model, DAG-F1 grows during the long reasoning phase before correctness increases, suggesting that the probe can capture useful intermediate reasoning structure before the final answer appears.
These patterns show that DAG recoverability provides a structural view of answer formation.

\begin{figure}[t]
\centering
\input{fig/temporal_generation_scaling_combined}
\caption{
\textbf{(a)} Qwen3-14B best-layer DAG-F1 for correct and incorrect final answers, across generated-token cutoffs; dashed lines denote appended-node probing, solid lines denote native-span probing, and shaded bands denote standard error.
\textbf{(b)} Qwen3 4B variants on ProofWriter, showing appended-node DAG-F1, native-span DAG-F1, and cumulative correctness across token cutoffs.
\textbf{(c)} Generation-time scaling of DAG recoverability across Qwen3 model sizes and non-Qwen models.}
\label{fig:temporal_generation}
\label{fig:scaling}
\end{figure}

\subsection{Generation-Time Recoverability Across Scale and Model Family}
We next ask whether the generation-time emergence pattern persists across model scale and family.
We apply the same generation-time analysis to a Qwen3 size sweep and several non-Qwen models.
Figure~\ref{fig:scaling}(c) shows a clear positive relationship between Qwen3 model size and DAG recoverability.
Qwen3-0.6B remains low and nearly flat across generation, and Qwen3-1.7B shows only a modest increase.
In contrast, the 4B, 8B, and 14B models start from higher DAG-F1 and continue to improve as more tokens are generated, with the ordering by size largely preserved throughout generation.
Thus, larger Qwen3 models show both higher generation-time DAG recoverability and more sustained gains as the reasoning trace advances.

The same panel shows that the qualitative behavior appears outside the Qwen3 family.
Among K2-Think \citep{cheng2025k2thinkparameterefficientreasoning}, Phi-4-reasoning \citep{abdin2025phi}, DeepSeek-R1-Distill-Llama-8B \citep{guo2025deepseek}, and Llama-3.2-3B \citep{meta_llama32_2024}, larger models show stronger generation-time DAG recoverability.
Together, these results point to a positive association between model capacity and DAG recoverability.

\subsection{Generation-Time DAG Steering}
\label{sec:dag_steering}
We next ask a more causal question: whether the DAG structure can be directly intervened on to change the model's continuation.
Using the DAG steering procedure described in Section~\ref{sec:steering_method}, we evaluate single-cutoff and multi-cutoff interventions.
For each example, the steered and no-steering continuations share the same first 384 prefix tokens, isolating the effect of the activation intervention on the subsequent continuation.

Table~\ref{tab:steering_dev} reports the resulting generation accuracy over three stochastic replay seeds.
On examples whose original generations were incorrect, DAG steering increases total correct continuations from 47 to 63 with an intervention at 384 tokens (+34.0\%), and to 71 and 73 with additional later interventions (+51.1 and 55.3\%).
On originally correct examples, the same settings leave accuracy nearly unchanged, with totals between 331 and 336 relative to a 332 no-steering baseline (-0.3\% to +1.2\%).
Thus, settings selected to recover failed generations do not come at the cost of degraded performance on generations that were already correct.
The improvements introduced by steering on the originally incorrect traces support the interpretation that the probed DAG structure is behaviorally relevant during reasoning, not only decodable after generation.

\begin{table*}[t]
\caption{Generation-time DAG steering on originally incorrect and correct ProofWriter examples.
All rows use Qwen3-4B at layer 21 with three stochastic replays over 121 examples.
Relative change is computed against the 384-token no-steering baseline within each group.}
\centering
\small
\newcommand{\steerpos}[1]{\textcolor{green!45!black}{#1}}
\newcommand{\steerneg}[1]{\textcolor{red!60!black}{#1}}
\newcommand{\steerneutral}[1]{\textcolor{gray}{#1}}
\setlength{\tabcolsep}{5pt}
\resizebox{\textwidth}{!}{%
\begin{tabular}{llllccc}
\toprule
Group & Condition & Cutoffs & Perturbation norm cap(s) & Per-seed correct & Total correct & Relative change \\
\midrule
\multirow{4}{*}{\begin{tabular}{@{}l@{}}Originally\\incorrect\\generations\end{tabular}} & No steering & 384 & -- & 15 / 17 / 15 & 47 / 363 & \steerneutral{--} \\
& Steered & 384 & 0.001 & 21 / 19 / 23 & 63 / 363 & \steerpos{+34.0\%} \\
& Steered & 384+1024 & 0.001, 0.0005 & 23 / 26 / 22 & 71 / 363 & \steerpos{+51.1\%} \\
& Steered & 384+1024+2048 & 0.001, 0.0005, 0.001 & 24 / 25 / 24 & 73 / 363 & \steerpos{+55.3\%} \\
\midrule
\multirow{4}{*}{\begin{tabular}{@{}l@{}}Originally\\correct\\generations\end{tabular}} & No steering & 384 & -- & 110 / 109 / 113 & 332 / 363 & \steerneutral{--} \\
& Steered & 384 & 0.001 & 110 / 107 / 114 & 331 / 363 & \steerneg{-0.3\%} \\
& Steered & 384+1024 & 0.001, 0.0005 & 115 / 111 / 110 & 336 / 363 & \steerpos{+1.2\%} \\
& Steered & 384+1024+2048 & 0.001, 0.0005, 0.001 & 116 / 109 / 109 & 334 / 363 & \steerpos{+0.6\%} \\
\bottomrule
\end{tabular}
}
\label{tab:steering_dev}
\end{table*}

\section{Discussion}
\paragraph{Reasoning as progressive graph construction.}
Our experiments provide three levels of evidence: 
static probing shows that DAG properties are linearly recoverable from hidden states; 
generation-time probing shows that this structure emerges during the model's own reasoning process; 
and steering shows that strengthening the probed structure can improve subsequent behavior.
Our results support a view in which reasoning progressively forms structured dependencies in representation space, rather than being captured solely by a linear trace.
This provides a complementary perspective to rationale-based analyses that does not depend on any particular verbalization \citep{elazar-etal-2021-amnesic}.

\paragraph{Localization of reasoning structure.}
Reasoning-relevant structure is not uniformly distributed across the network.
DAG properties are most recoverable in intermediate layers, consistent with layer specialization \citep{tenney2019bertpipeline} and with the existence of reasoning-dominant layers \citep{he2025far}.
The depth- and span-dependent patterns further suggest that node position and long-range dependency information emerge at different stages of the network, pointing to partial localization of distinct structural reasoning components useful for future mechanistic analysis or intervention.

\paragraph{Relation to chain-of-thought.}
A linear rationale can be viewed as a projection of an underlying dependency structure, 
so multiple faithful traces may correspond to the same internal graph 
and faithful reasoning need not have a unique textual explanation.
Conversely, generated CoT can be fluent yet unfaithful or lead to wrong answers \citep{turpin2023language, lanham2023measuring}, making textual rationales alone an incomplete view of the underlying reasoning process.
Probing DAG structure therefore complements chain-of-thought analysis by focusing on structural properties that can persist across different verbalizations.

\paragraph{Implications for model design.}
Structured dependency information may provide a target for model design, training, and control.
Architectures or objectives that preserve or expose latent graph structure could improve interpretability, while interventions or decoding strategies that better align generation with structured internal representations may improve reasoning reliability \citep{chen2026chains, besta2024graph}.

\section{Conclusion}
We introduced \emph{Reasoning DAG Probing}, a framework for examining whether and where large language models encode graph-structured reasoning in their internal representations.
By formalizing multi-step reasoning as a directed acyclic graph and probing for node depth, pairwise distance, and pairwise adjacency, we showed that key properties of reasoning structure are linearly recoverable from hidden states.
This structure is most accessible in intermediate layers, aligns with reasoning depth and edge span, increases with model size, and emerges during generation both when the model is queried about specific reasoning states and within its own generated reasoning trace.

Our results suggest that LLM reasoning is not well characterized as a purely linear process.
Instead, internal representations reflect structured dependencies that extend beyond surface chain-of-thought traces and are only partially expressed through text.
Generation-time steering further shows that strengthening the model's internal organization toward the correct reasoning graph can recover some originally incorrect generations while largely preserving already-correct ones, suggesting that the probed structure is behaviorally relevant rather than only decodable.

Overall, these findings position reasoning DAGs as a useful abstraction for mechanistic analysis.
By shifting attention from linear explanations to structural properties of internal representations, our work takes a step toward a more faithful characterization of how multi-step reasoning is represented and computed in large language models.

\section*{AI Use Statement}
In this work, we used generative AI tools to construct the GSM8K reasoning DAG annotations described in Appendix~\ref{sec:gsm8k}, where GPT-5 mini was prompted to extract structured reasoning steps from each example's question and worked answer.
We also used AI-assisted tools for language editing, phrasing, and clarity improvements.
We have not used generative AI tools for retrieval and discovery, research ideation or execution, or to make or prove scientific claims.
All AI-assisted annotations and text edits were reviewed by the authors; the resulting DAGs were filtered and evaluated using the procedures described in the paper.
We take responsibility for the final content of this work, including text, data annotations, and artifacts produced with the aid of generative AI.

\bibliographystyle{iclr2027_conference}
\bibliography{iclr2027_conference}

\clearpage

\appendix

\section{Limitations}
Our study focuses on datasets with explicit proof structure, providing a controlled setting for analyzing reasoning representations, though extending this framework to more open-ended and naturalistic reasoning remains an important direction for future work.
In addition, our evaluation targets depth, distance, and adjacency as interpretable summaries of reasoning structure; richer annotations or alternative formalisms could further broaden the range of properties that can be probed, such as rule types or edge semantics.
Finally, although DAG structure is not directly observable at test time for arbitrary user queries, our findings point to promising opportunities for leveraging internal structure, for example through structured prompting, auxiliary supervision, or representation-level interventions.

\section{Related Work}
\label{sec:related_work}
\paragraph{Chain-of-thought and faithfulness.}
CoT prompting has become a widely used technique for improving performance on multi-step reasoning tasks \citep{wei2022chain, zelikman2022star, prystawski2023think, wang2022iteratively, liu2023crystal, feng2023towards}.
At the same time, there is ongoing debate about whether generated rationales are faithful to the underlying computation and how to evaluate faithfulness \citep{turpin2023language,lanham2023measuring, kadavath2022language}.
Our work is motivated by this debate and focuses on structure inside activations instead of surface explanations.

\paragraph{Probing and structural probes.}
Probing methods aim to measure what information is encoded in neural representations by training constrained predictors on frozen features, with ongoing discussion of their promises and limitations \citep{belinkov2019analysis,belinkov2022probing}.
Within this framework, \citet{hewitt-manning-2019-structural} introduce the structural probe, showing that a low-rank linear map can recover structured objects, such as syntactic tree distances, from transformer representations.
Subsequent work has extended structural probes and related techniques to a variety of linguistic and conceptual structures \citep{wang2022interpretability, liu2025predict, jing2025lingualens, madusanka2023not, zahid2024probing}.
We build directly on this methodology, but replace syntactic trees with reasoning DAGs, and replace parse-tree distance with graph depth and pairwise distance in the reasoning structure.

\paragraph{Minimal-pair probing and contrastive evaluation.}
A complementary line of work studies representational structure through \emph{minimal pairs}, where inputs differ by a controlled change and the analysis emphasizes the induced contrast rather than absolute accuracy \citep{misra2023comps, he2025xcomps}.
Minimal-pair benchmarks have been widely used for targeted evaluation, and more recent work formalizes minimal-pair probing as a decoding problem to localize where particular distinctions become linearly accessible in a model's representation space \citep{he2024decoding}.
This style of controlled, contrastive analysis is closely related to our goal of isolating structural signals while controlling for superficial cues.
In particular, \citet{he2025far} investigate representations for reasoning and propose minimal-pair based analyses that emphasize carefully controlled contrasts; we adopt a similar philosophy when designing controls that break graph structure while preserving surface form.

\paragraph{Graph-structured views of reasoning.}
Representing reasoning as a graph, such as entailment graphs, proof graphs, or dependency DAGs over intermediate conclusions, has been explored in natural language inference, multi-hop question answering, and formal proof settings \citep{saparov2022language}.
Graph representations make explicit which intermediate results depend on which premises, enabling reuse and branching that linear chain-of-thought cannot express without redundancy.
Recent work on graph-based prompting and reasoning further highlights the distinction between internal reasoning structure and its textual linearization \citep{yao2023tree,besta2024graph}.
Our work complements these approaches by introducing a probing-based measurement framework that directly tests whether such graph structure is encoded in LLM hidden states.

\paragraph{Measuring reasoning structure in LLMs.}
There is increasing interest in identifying internal circuits and representations underlying reasoning in LLMs, including layerwise analyses, representation similarity methods, and causal interventions \citep{hong2024aimpliesb, ameisen2025circuittracing, burns2022discovering, wang2024grokked, hanna2023does}.
Compared to purely causal or purely behavioral evaluations \citep{kassner2020pretrained, wang2024understanding}, probing offers a lightweight and scalable way to test whether key structural variables are accessible from hidden states.
By focusing on DAG geometry, our study complements prior work that examines linear CoT sequences, and provides a way to ask whether the model internally organizes intermediate states into a coherent dependency structure.

\section{Example ProofWriter Question}
\label{sec:proofwriter_example}
We illustrate our DAG formulation using a concrete example from ProofWriter.
The following theory--query pair is drawn from the dataset and admits a non-trivial proof structure.

\begin{figure*}[t]
\centering
\includegraphics[width=0.7\textwidth]{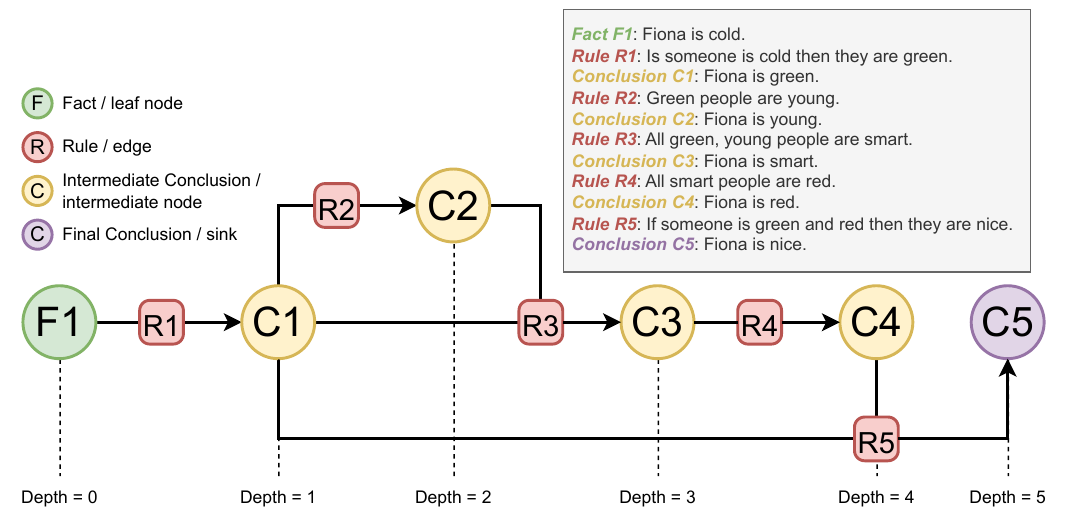}
\caption{Reasoning DAG for a ProofWriter example.
Nodes represent premises, intermediate conclusions, and the final answer; edges correspond to rule applications.
The graph illustrates both the need for longest-path depth to capture premise participation and the presence of valid dependencies spanning multiple inference steps. }
\label{fig:dag_example}
\end{figure*}

\paragraph{Theory.}
\small
\begin{quote}
Dave is cold.
Dave is smart.
Dave is green.
Dave is young.
Erin is cold.
Erin is kind.
Erin is red.
Erin is smart.
Erin is green.
Fiona is cold.
Fiona is kind.
Gary is kind.
Gary is red.
Gary is smart.

If someone is green and red then they are nice.
All smart people are red.
If someone is kind and smart then they are cold.
Nice people are smart.
If someone is cold then they are green.
Green people are young.
All green, young people are smart.
Nice people are green.
If Gary is kind and Gary is green then Gary is smart.
\end{quote}
\normalsize
\paragraph{Query.}
\small
\begin{quote}
Fiona is nice.
\end{quote}
\normalsize
Figure~\ref{fig:dag_example} shows the reasoning DAG constructed from the gold proof for this example.
Nodes correspond to factual statements or intermediate conclusions, while directed edges represent rule applications linking premises to derived conclusions.
The final answer node (\emph{Fiona is nice}) appears as the designated sink of the graph.

\paragraph{Why longest-path depth.}
\label{paragraph:why_longest}
Figure~\ref{fig:dag_example} illustrates why node depth is defined by the longest directed path to the sink.
Consider node $C1$, corresponding to \emph{Fiona is green} (second from the left).
Although $C1$ directly supports the final conclusion $C5$ \emph{Fiona is nice} via rule $R5$ \emph{green \& red $\rightarrow$ nice}, it also participates in a longer chain: \emph{green $\rightarrow$ young}, \emph{green \& young $\rightarrow$ smart}, \emph{smart $\rightarrow$ red}, and finally \emph{green \& red $\rightarrow$ nice}.
Thus, $C1$ lies multiple inference steps upstream of the final answer even though a shorter path exists through a rule application that $C1$ alone does not satisfy.
Shortest-path depth would collapse these roles and underestimate a premise's contribution to the full reasoning process.
By contrast, longest-path depth assigns greater depth to premises that participate in extended chains, especially when multiple premises jointly support a conclusion through a single rule.

\paragraph{Why long-range edges matter.}
\label{paragraph:long_edge}
The same example also shows why dependency edges spanning multiple inference steps must be allowed.
In Figure~\ref{fig:dag_example}, the final conclusion $C5$ \emph{Fiona is nice} depends on $C1$ \emph{Fiona is green}, which lies several steps upstream in the DAG.
Restricting reconstruction to only local or adjacent-depth edges would therefore discard such long-range but structurally meaningful connections.
This motivates using moderate-to-large distance thresholds and non-trivial depth gaps to recover valid edges linking nodes across multiple inference layers.

\section{Different Model Families}
\label{sec:different_model}
Beyond the Qwen3 family, we evaluate reasoning DAG recoverability across several additional LLM families that differ in model scale, architecture, and training objectives.
Specifically, we analyze K2-Think (32B) \citep{cheng2025k2thinkparameterefficientreasoning}, Phi-4-reasoning \citep{abdin2025phi}, DeepSeek-R1-Distill-Llama-8B \citep{guo2025deepseek}, and Llama-3.2-3B \citep{meta_llama32_2024}.
Despite these differences, we consistently observe that reasoning DAG structure is linearly recoverable from intermediate hidden states across all models considered.

\begin{figure*}[t]
\centering
\includegraphics[width=0.48\textwidth]{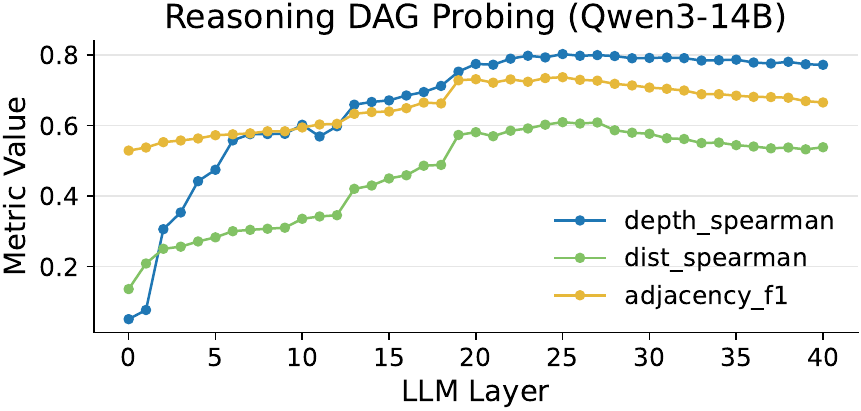}
\includegraphics[width=0.48\textwidth]{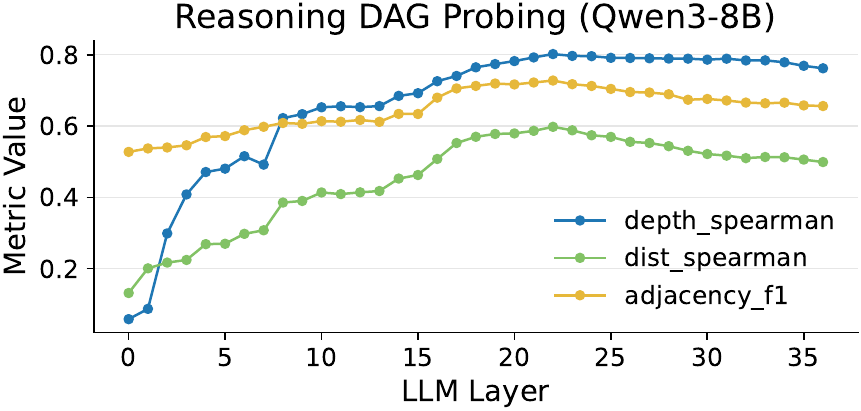}
\includegraphics[width=0.48\textwidth]{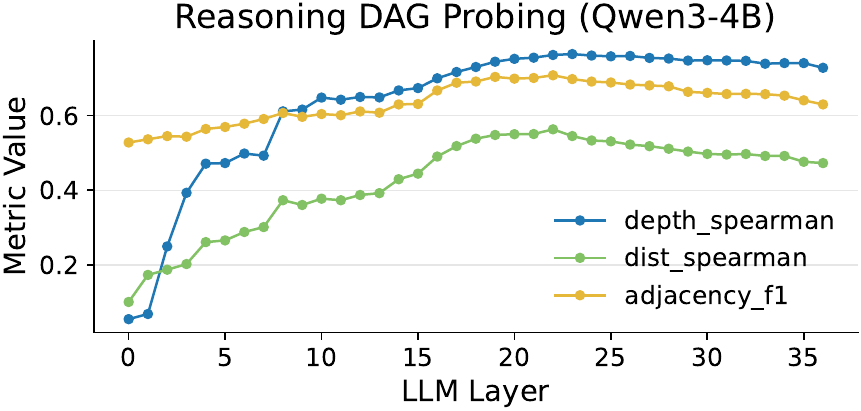}
\includegraphics[width=0.48\textwidth]{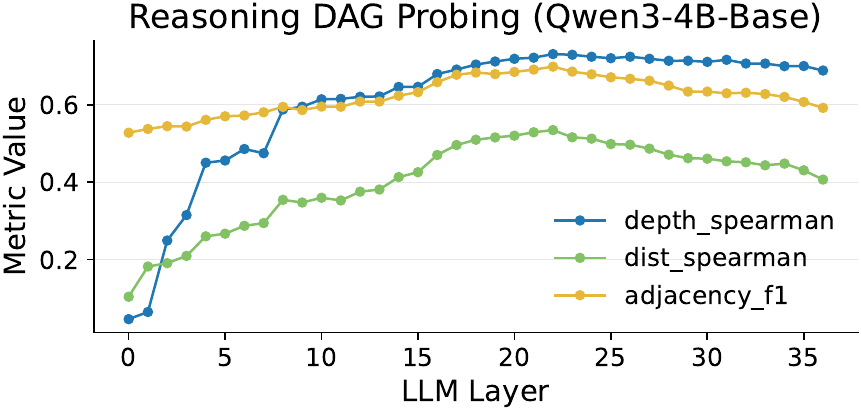}
\includegraphics[width=0.48\textwidth]{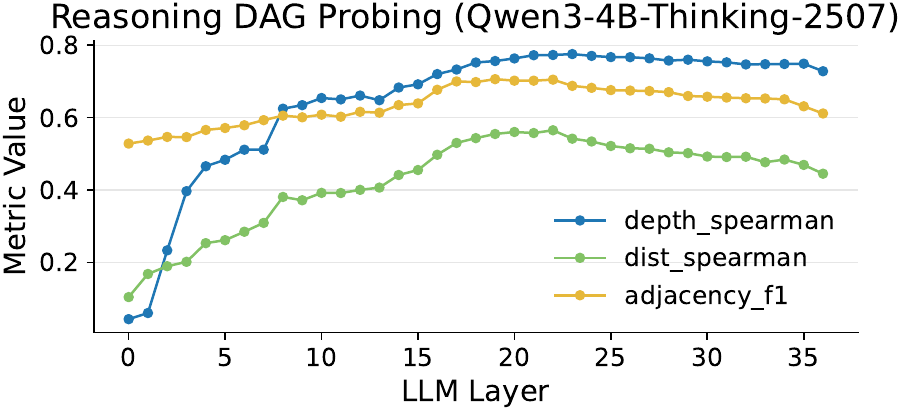}
\includegraphics[width=0.48\textwidth]{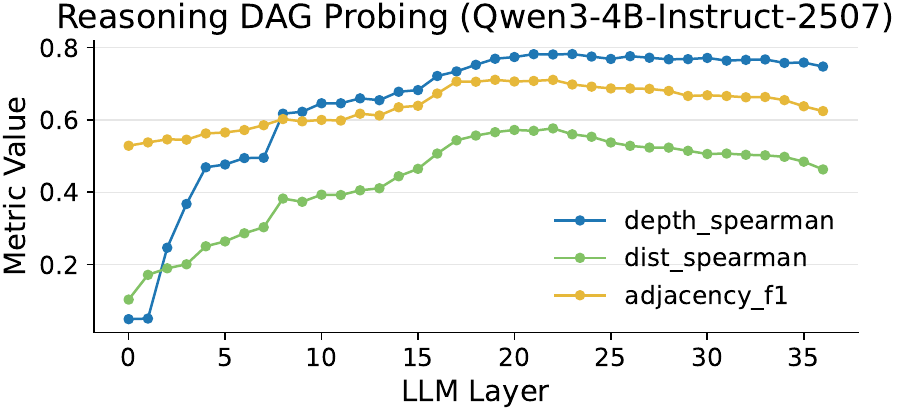}
\includegraphics[width=0.48\textwidth]{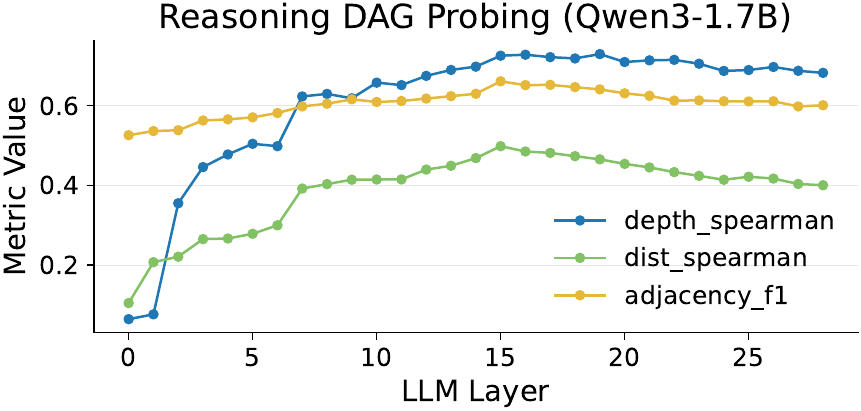}
\includegraphics[width=0.48\textwidth]{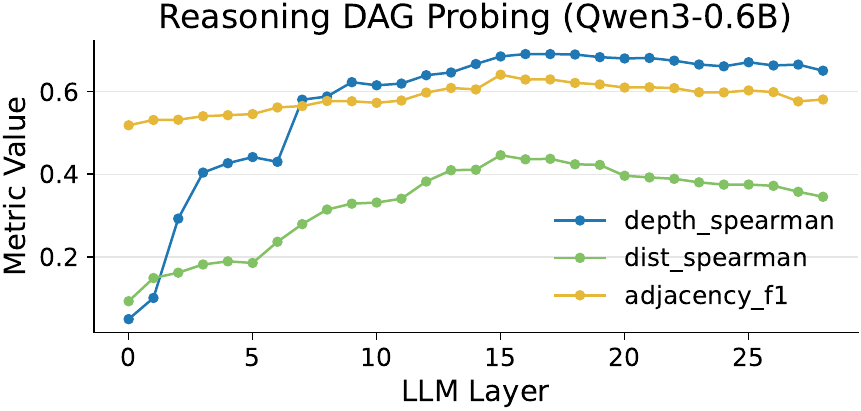}
\caption{Layerwise probing performance for reasoning DAG recoverability across the Qwen3 model family.
Each panel reports depth Spearman correlation, distance Spearman correlation, and adjacency F1 as a function of layer depth.
Across model scales and training variants, DAG structure is most strongly recoverable in intermediate layers. }
\label{fig:layerwise_qwen}
\end{figure*}

\begin{figure*}[t]
\centering
\includegraphics[width=0.48\textwidth]{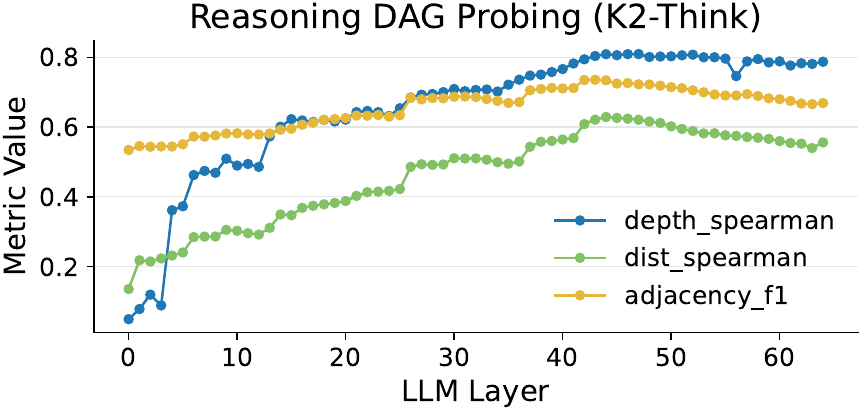}
\includegraphics[width=0.48\textwidth]{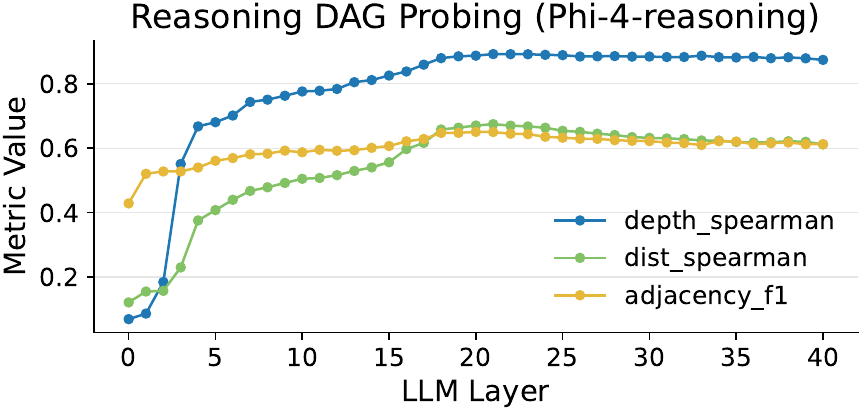}
\includegraphics[width=0.48\textwidth]{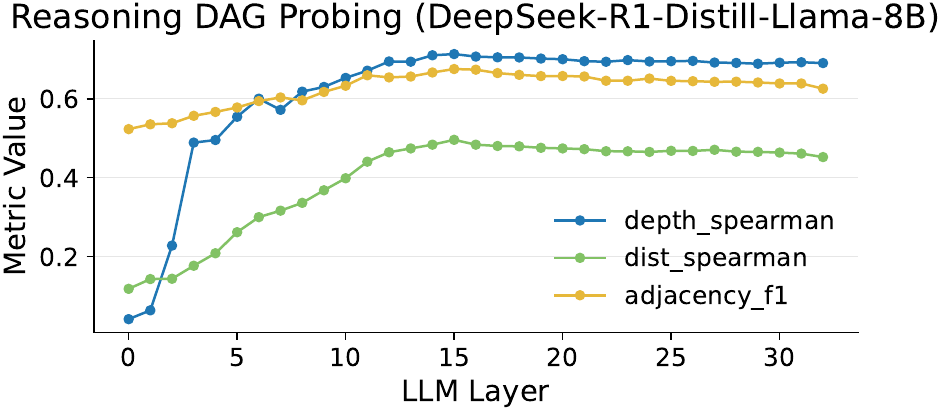}
\includegraphics[width=0.48\textwidth]{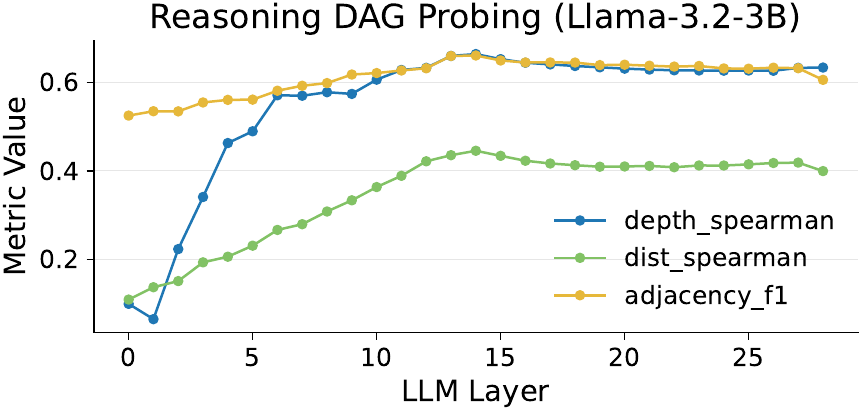}
\caption{Layerwise probing performance for reasoning DAG recoverability across non-Qwen model families.
Despite differences in architecture, scale, and training objectives, all models exhibit a similar qualitative pattern, with reasoning DAG geometry most accessible in intermediate layers. }
\label{fig:layerwise_nonqwen}
\end{figure*}

Figures~\ref{fig:layerwise_qwen} and~\ref{fig:layerwise_nonqwen} summarize the layerwise probing results for all evaluated models.
Across families, recoverability exhibits a broadly similar qualitative pattern: probing performance improves from early layers, peaks in intermediate layers, and becomes more variable in later layers.
While absolute performance varies with model size and training recipe, the presence of a reasoning-dominant band of layers appears to be a robust property shared across architectures.

\section{Alternative Training Objectives for Depth Probing}
In the main experiments, depth probes are trained using a pairwise ranking objective that enforces local ordering constraints between adjacent depth levels.
To assess the robustness of our findings to the choice of supervision signal, we additionally experiment with regression- and classification-based training objectives for depth probing.

\paragraph{Regression objective.}
We retain the same probe architecture as in the ranking setup: a rank-1 linear map without bias,
\[
\hat{\operatorname{depth}}(v) = \mathbf{w}^\top \mathbf{z}_v ,
\]
where $\mathbf{z}_v$ is the pooled hidden representation of node $v$ at a given layer.
The probe is trained to predict the scalar gold depth $\operatorname{depth}(v)$ using mean-squared error,
\[
\mathcal{L}_{\text{reg}} = \bigl(\hat{\operatorname{depth}}(v) - \operatorname{depth}(v)\bigr)^2 ,
\]
optimized with AdamW.
Model selection is performed based on development loss, and evaluation follows the same protocol as in the ranking setting.

\paragraph{Classification objective.}
For classification, we instead train a linear classifier over discrete depth labels.
Let $K = 5$ denote the maximum depth considered; we restrict supervision to nodes with depths in $\{0,\dots,5\}$ and define
\[
\mathbf{p}_v = \mathrm{softmax}(\mathbf{w}^\top \mathbf{z}_v),
\]
where $W \in \mathbb{R}^{K \times d}$.
The probe is trained with cross-entropy loss against the gold depth class, and test-time predictions are obtained via $\arg\max_k \mathbf{p}_{v,k}$.
All other aspects of the setup, including data splits and depth annotations, are identical to the ranking and regression variants.

\begin{figure*}[t]
\centering
\includegraphics[width=0.75\textwidth]{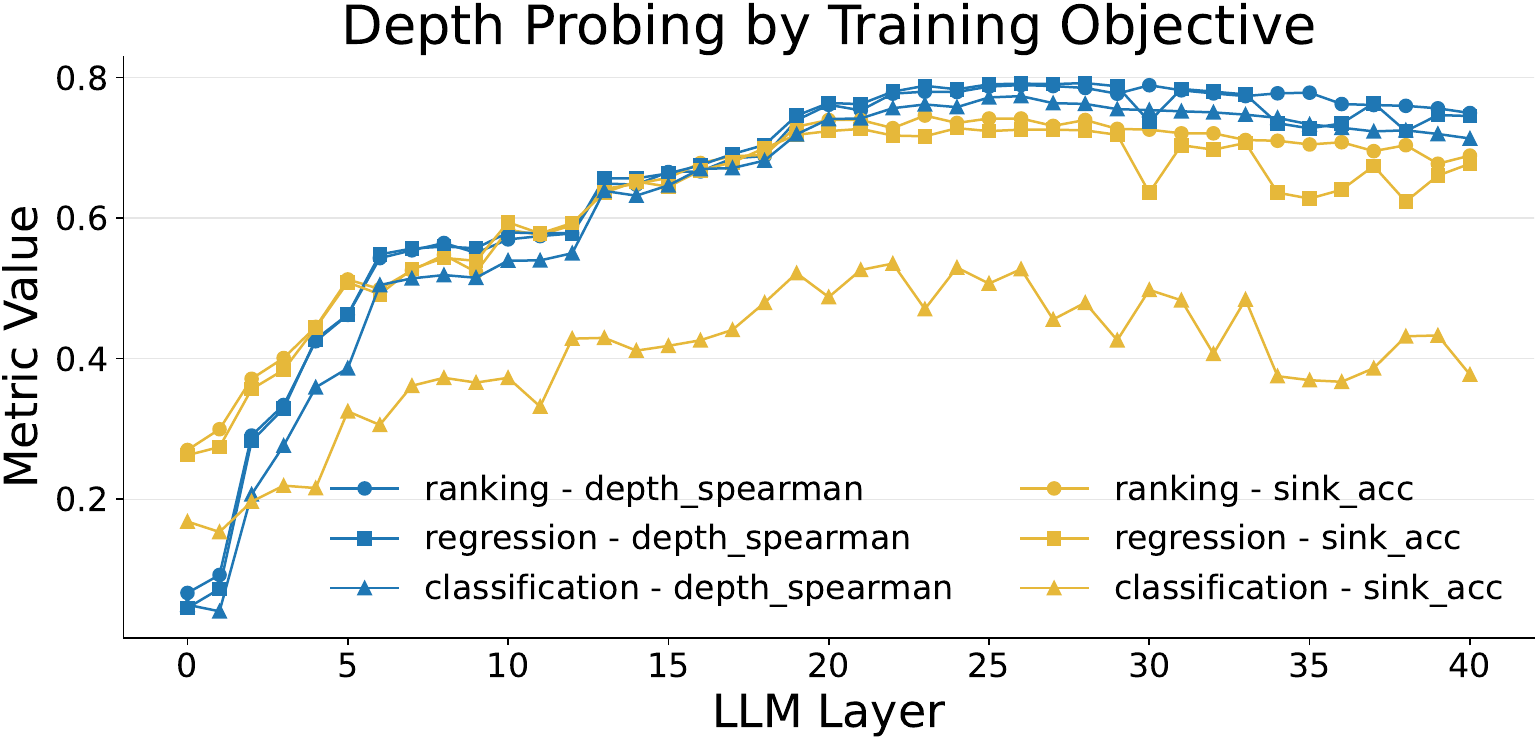}
\caption{Layerwise depth probing performance on Qwen3-14B under different training objectives.
We report depth Spearman correlation and sink accuracy for ranking-, regression-, and classification-based depth probes.
Ranking and regression objectives yield similar qualitative trends and strong intermediate-layer performance, while classification exhibits weaker sink identification and higher variability. }
\label{fig:compare_training}
\end{figure*}

\paragraph{Comparison across objectives.}
Figure~\ref{fig:compare_training} compares layerwise depth Spearman correlation and sink accuracy for the three training objectives on Qwen3-14B.
All objectives recover a broadly similar qualitative pattern, with depth information becoming increasingly accessible in early layers and peaking in intermediate layers.
Both ranking and regression objectives achieve strong depth ordering performance across layers.
The ranking objective in particular exhibits more stable behavior in the final layers, motivating its use as the primary training objective in our main experiments.
In contrast, the classification objective consistently underperforms on sink accuracy and exhibits greater variability across layers, suggesting that discretizing depth into classes discards useful ordinal structure.
Overall, these results indicate that the emergence of depth information is robust to the choice of continuous training objective, while objectives that fail to respect depth ordering are less effective at recovering global reasoning structure.

\section{Procedural DAG Reconstruction Case Study}
\begin{figure*}[t]
\centering
\includegraphics[width=\textwidth]{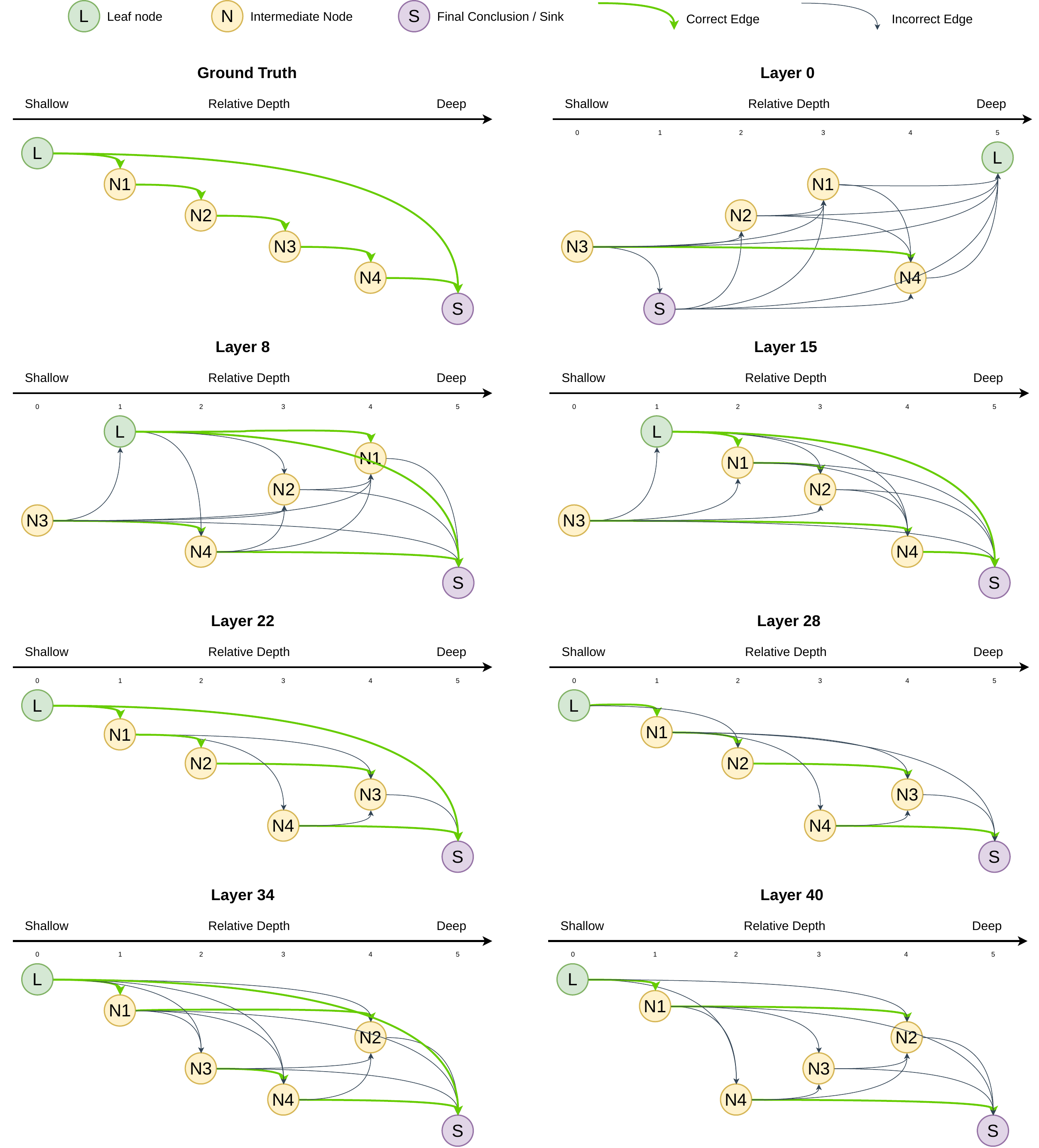}
\caption{Procedural DAG reconstruction across layers for a representative test example.
Nodes are ordered from top to bottom by ground-truth depth (shallow to deep) and from left to right by predicted relative depth (shallow to deep).
Green arrows indicate correctly recovered gold edges, while charcoal arrows denote incorrect predicted edges.
Early layers exhibit unstable depth ordering and noisy connectivity; intermediate layers recover both correct node ordering and dependencies; later layers retain coarse structure but show a mild decline in edge-level accuracy. }
\label{fig:case_study}
\end{figure*}

We present a qualitative case study of how reasoning DAG structure emerges across layers for a representative test example.
Figure~\ref{fig:case_study} shows reconstructed DAGs at selected layers of Qwen3-14B.
Nodes are arranged from top to bottom by ground-truth depth (shallow to deep) and from left to right by predicted depth (shallow to deep).
Green arrows denote predicted edges that match gold DAG edges, while charcoal arrows denote predicted edges absent from the gold structure.

At layer~0, both relative depth ordering and edge directions are highly unstable: leaf, sink, and intermediate nodes are interleaved, and many edges are misdirected or spurious, indicating that early representations lack coherent global structure.
By layer~8, a coarse depth ordering begins to emerge: the leaf node is placed shallower than three of the intermediate nodes, with the sink at the deepest end.
Up to four gold edges are recovered, but this appears to reflect weak discrimination rather than precise structural encoding.
Nodes receive similar predicted depths, producing dense edge predictions with high coverage but low precision and many incorrect long-range connections.

By layer~15, the predicted node ordering becomes substantially cleaner, with three intermediate nodes correctly positioned relative to each other.
One additional correct dependency is recovered and one spurious edge is removed, though many spurious edges remain, implying high recall but low precision.
At layer~22, reconstruction quality peaks: five of six gold edges are recovered, relative node depths are nearly perfectly ordered, and incorrect edges are sparse.
This closely matches the peak reconstruction and probing performance observed quantitatively in Figure~\ref{fig:probe_reconstruct}(a).

Beyond this point, recoverability gradually declines.
At layer~28, node ordering is unchanged and the overall graph shape remains similar, but the longest gold edge is no longer recovered and several incorrect edges reappear.
Layers~34 and~40 show similar structures but with more spurious edges, indicating not an abrupt degradation but a smooth, mild loss of structural fidelity.
This suggests that later layers preserve the coarse reasoning scaffold while becoming less selective in encoding precise dependency relations.

Overall, this case study visually confirms our quantitative findings: reasoning DAG structure emerges progressively, is represented most faithfully in intermediate layers, and shows a gradual, mild loss of precision in later layers instead of a sharp collapse.

\section{Layerwise Performance of Baselines}
We report layerwise probing results for two baselines: \emph{node-only} and \emph{label-shuffled}.
Because bag-of-words features are not layer-indexed, the \emph{bag-of-words} baseline is instead summarized in Figure~\ref{fig:probe_reconstruct}(b).

\begin{figure*}[t]
\centering
\includegraphics[width=0.75\textwidth]{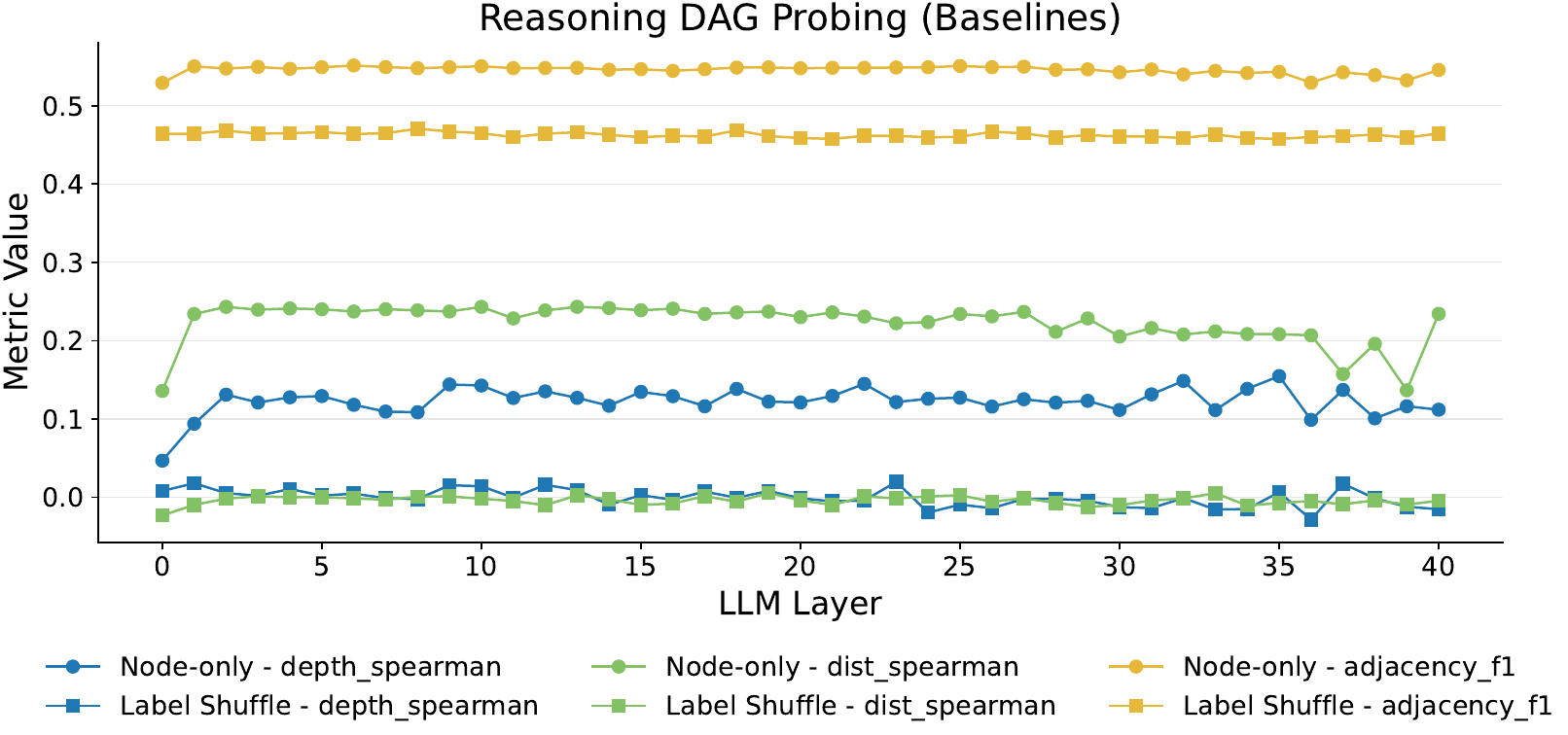}
\caption{Layerwise probing performance for node-only and label-shuffled baselines.
We report depth Spearman correlation, distance Spearman correlation, and adjacency F1 across layers.
Node-only representations exhibit weak and largely flat performance, while label shuffling collapses all metrics toward chance, indicating that recoverable DAG structure depends on aligned contextual representations rather than probe expressivity. }
\label{fig:layerwise_baselines}
\end{figure*}

Figure~\ref{fig:layerwise_baselines} shows that both baselines fail to recover meaningful reasoning DAG structure across layers.
For the node-only baseline, depth, distance, and adjacency metric values remain low and largely flat with little systematic layerwise improvement, indicating that isolated node text provides only weak, non-progressive signals that do not benefit from deeper contextual processing or reasoning.

The label-shuffled baseline collapses almost entirely across all metrics.
Depth and distance correlations remain near zero at all layers, and adjacency F1 is close to the performance of a random decoder.
This confirms that probe performance in the main method reflects alignment between hidden representations and the reasoning DAG, showing that recoverable DAG geometry arises from contextualized representations integrating the full theory, not from probe capacity or shallow textual features.

\section{Reconstruction Metrics Across Layers}
We evaluate reasoning DAG reconstruction quality across layers for Qwen3-14B, reporting edge-level precision, recall, and F1 using the best-performing distance cutoff $\tau_{\text{dist}}$ and adjacency threshold $\tau_{\text{adj}}$ at each layer.

\begin{figure*}[t]
\centering
\includegraphics[width=0.75\textwidth]{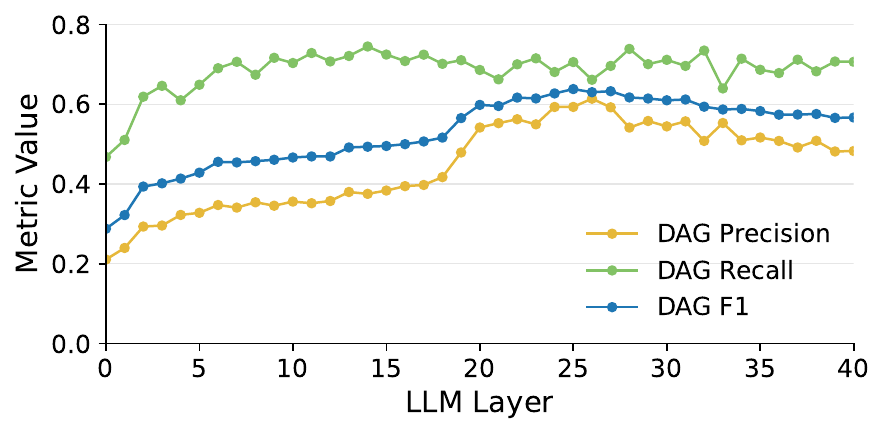}
\caption{Layerwise DAG reconstruction performance for Qwen3-14B.
We report edge precision, recall, and F1 at each layer, selecting the best reconstruction thresholds over $\tau_{\text{dist}}$ and $\tau_{\text{adj}}$.
Reconstruction quality improves rapidly in early layers and peaks in intermediate layers, with recall consistently exceeding precision, showing broad recovery of dependencies but reduced selectivity in later layers. }
\label{fig:layerwise_recon}
\end{figure*}

Figure~\ref{fig:layerwise_recon} shows a clear layerwise pattern: reconstruction improves rapidly in early layers, indicating that even shallow representations contain partial structural information.
Recall rises quickly and remains consistently above precision, suggesting that the reconstruction procedure recovers many true dependencies but also introduces spurious edges.
Precision increases more gradually and peaks in intermediate layers, reflecting improved discrimination of valid dependencies as representations become more structured, then declines slightly in later layers as selectivity weakens, consistent with the pattern in Figure~\ref{fig:case_study}.
F1 likewise reaches a broad maximum in intermediate layers, mirroring the trends observed for depth, distance, and adjacency probing in Figure~\ref{fig:probe_reconstruct}(a).

\section{Group-wise Normalization for Adjacency Recoverability}
\label{app:adj_norm}
For the span-dependent adjacency analysis in Figure~\ref{fig:depth_binned_mae}(b), performance is normalized separately within groups defined by depth gap.
Because adjacency prediction is a binary task and we evaluate it using average precision (AP), absolute scores are strongly affected by the positive-class rate in each group.
In particular, the baseline AP of a random classifier roughly equals the prevalence of the positive class, so groups with different class ratios are not directly comparable in raw AP.

This issue is substantial in our setting because the adjacency ratio varies sharply with depth gap.
In a directed acyclic graph, node pairs with depth gap 1 are often directly connected, since adjacent reasoning depths frequently correspond to parent-child relations.
By contrast, for larger depth gaps, most node pairs are not adjacent: although a direct edge spanning multiple reasoning steps is still possible (Appendix~\ref{paragraph:long_edge}), it becomes progressively less likely.
Table~\ref{tab:depth_gap_stats} shows that the positive class rate decreases markedly as depth gap increases in the test set.

\begin{table}[t]
\caption{Class statistics for node adjacency grouped by depth gap on the test set.}
\centering
\small
\begin{tabular}{rcccc}
\toprule
Depth gap & $n$ & Pos & Neg & Pos class rate \\
\midrule
1 & 6,026 & 5,425 & 601  & 0.9003 \\
2 & 4,816 & 518  & 4,298 & 0.1076 \\
3 & 3,575 & 365  & 3,210 & 0.1021 \\
4 & 2,342 & 202  & 2,140 & 0.0863 \\
5 & 1,145 & 88   & 1,057 & 0.0769 \\
\bottomrule
\end{tabular}
\label{tab:depth_gap_stats}
\end{table}

As a result, raw AP values from different depth-gap groups lie on different scales and cannot be meaningfully compared directly.
To make cross-group trends visually comparable and to better show how recoverability evolves across layers, we apply min-max normalization within each group before plotting:
\[
\tilde{x}_{g,\ell} = \frac{x_{g,\ell} - \min_{\ell'} x_{g,\ell'}}{\max_{\ell'} x_{g,\ell'} - \min_{\ell'} x_{g,\ell'}} ,
\]
where $x_{g,\ell}$ denotes the AP for depth-gap group $g$ at layer $\ell$, and $\tilde{x}_{g,\ell}$ is the normalized value.

\section{Generation-Time DAG Recoverability Details}
\label{app:dag_gen_details}
\paragraph{Trace generation.}
For each selected example, we construct a task-specific prompt from the input instance and ask the model to produce a reasoning trace followed by a final answer in the benchmark's answer format.
The prompt is adapted to the benchmark, but keeps the same high-level structure: identify the task, request step-by-step reasoning, specify the final-answer format, and provide the problem input.
For chat-format checkpoints, this user prompt is wrapped with the model's tokenizer chat template before generation.
Figure~\ref{fig:generation-prompt-template} shows the prompt schema used for generation-time evaluation.
We generate continuations up to a fixed token budget and record the generated text, generated token ids, and parsed final answer.

\begin{figure*}[t]
\centering
\small
\begin{promptbox}[unbreakable,title={Generation Prompt Template}]
\ttfamily
\setlength{\parskip}{4pt}
You are given a \{task description\}.

\{Task-specific instruction\}

Think step by step, then finish with:

\{task-specific final-answer format\}

\psection{Problem:}
\{problem input\}

\psection{Question:}
\{question, if applicable\}
\end{promptbox}
\caption{Prompt schema used for generation-time evaluation.
For ProofWriter, the final answer is \texttt{<answer> true </answer>} or \texttt{<answer> false </answer>}.
For GSM8K, it is \texttt{<answer> final numeric answer </answer>}.
For TACO, it is a fenced Python code block containing the complete solution.}
\label{fig:generation-prompt-template}
\end{figure*}

\paragraph{Generation-time representations.}
To measure how recoverable DAG structure evolves during autoregressive generation, we evaluate hidden states after fixed generated-token cutoffs.
In our experiments, these cutoffs are 0, 32, 64, 128, 192, 256, 384, 512, 768, 1024, 1536, 2048, 3072, 4096, 6144, and 8192 tokens, where cutoff 0 corresponds to the model state before any generated answer tokens are appended.
For each cutoff \(c\), we encode the same task input together with the first \(c\) generated tokens and extract node representations for the corresponding reasoning DAG in two ways.

\emph{Appended-node probing} follows the static probing setup in Section~\ref{subsec:probing_setup}.
After the generated prefix, we append canonical text for each DAG node and use the hidden states on these appended node strings as node representations.
This gives a controlled readout of all graph nodes at every cutoff, making probe evaluation directly comparable across cutoffs even when the model has not yet explicitly mentioned every node.

\emph{Native-span probing} instead uses the model's own generated trace.
We align gold DAG nodes to their mentions in the generated text and, at cutoff \(c\), use the latest aligned span for each node that has appeared before the cutoff.
No canonical node text is appended in this setting: the probe reads only from tokens produced by the model itself.
This makes the representation less uniformly covered than appended-node probing, but provides a more direct measure of whether DAG structure is present in the generated reasoning trace.

\paragraph{Probing and evaluation.}
For appended-node probing, we train probes separately for each generated-token cutoff and model layer, since the appended node positions are controlled at each cutoff.
For native-span probing, node mentions occur at model-determined positions, so we train one probe per layer using native spans across token lengths and evaluate it at each cutoff.
We report DAG-F1 from reconstructed edges against gold graph edges.
Final-answer correctness is computed with the benchmark-specific answer parser and paired with the DAG-F1 trajectory for the same examples.
Figures~\ref{fig:temporal_generation} and~\ref{fig:temporal_4b_tasks} report these generation-time DAG-F1 and correctness trajectories.
\begin{figure*}[t]
\centering
\input{fig/temporal_4b_tasks}
\caption{Generation-time DAG recoverability and cumulative correctness across three generation tasks for Qwen3 4B variants.
For each task, the upper plot shows average DAG-F1 of the best probing layer at each generated-token cutoff, and the lower plot shows cumulative generation correctness.
The DAG-F1 trajectories closely anticipate the later correctness trajectories: they preserve the relative ordering of model variants and often show the same shape over generation, such as early spikes followed by plateaus or delayed rapid growth.
This alignment is strongest on ProofWriter, where explicit rule-chain structure makes recoverable DAG organization closely track answer emergence.
On GSM8K and TACO, the association remains visible despite additional noise: DAG-F1 still anticipates the broad correctness trends and preserves the eventual ordering of model variants, though the trajectories are less clean because arithmetic word problems and open-ended program synthesis express reasoning structure less directly than ProofWriter's explicit rule chains.}
\label{fig:temporal_4b_tasks}
\end{figure*}

\section{Generation-Time DAG Steering Details}
\label{app:dag_steering_details}
The steering experiments intervene on hidden states associated with native mentions of DAG nodes in the generated trace.
For a cutoff \(c\), we first align the prefix up to \(c\) generated tokens to the gold DAG and select the latest generated span for each observed node.
The intervention is applied at a fixed transformer layer to the residual-stream hidden states at these selected token positions.

To compute the steering direction, we attach a learnable additive perturbation to the selected hidden states and evaluate the depth, distance, and adjacency probes on the perturbed representations.
These are the shared native-span probes described in Appendix~\ref{app:dag_gen_details}: for each layer, a single probe is trained across spans occurring at different token positions and is reused for all steering cutoffs.
The perturbation is optimized to reduce the probe loss toward the gold DAG structure.
After only one gradient step, we normalize the gradient direction and project the update to a norm budget, expressed as a fraction of the Frobenius norm of the selected hidden states.
This keeps the intervention comparable across examples and cutoffs with different numbers of selected spans.

After the perturbation is chosen, we do not continue generation from an inconsistent cache.
Instead, we replay the prompt and generated prefix with teacher forcing, inject the perturbation at the selected layer and token positions during this replay, and allow all downstream layers and later prefix positions to be recomputed from the intervened states.
The key-value cache used for subsequent autoregressive continuation is therefore rebuilt from the steered prefix rather than patched locally.
For multi-cutoff steering, we repeat this procedure at each cutoff: spans are realigned within the current generated trace, a new perturbation is optimized for that cutoff, and the continuation proceeds from the newly rebuilt cache.

Hyperparameters are selected on development examples.
We tune the generated-token cutoffs, perturbation norm budgets, then report a small set of settings that were selected before evaluation on the held-out groups.
The reported main settings use Qwen3-4B layer 21, a single gradient step at each cutoff, equal depth/distance/adjacency loss weights, and norm budgets selected from the development sweeps.

\section{Generalization to GSM8K}
\label{sec:gsm8k}
\subsection{GSM8K as an Out-of-domain Testbed}
We further evaluate our probing framework on GSM8K \citep{cobbe2021gsm8k}, a benchmark of grade-school math word problems.
Unlike ProofWriter, which consists of rule-based natural language inference examples with explicitly structured symbolic proofs, GSM8K involves arithmetic reasoning grounded in natural language problem solving.
It therefore serves as an out-of-domain setting that differs both in task format and in the form of intermediate reasoning.

Each GSM8K example contains two key fields: a \texttt{question}, which presents the math word problem in natural language, and an \texttt{answer}, which includes a step-by-step solution followed by the final numerical answer.
We use these fields to derive structured reasoning graphs for probing.

An example is shown below:

\paragraph{Question.}
\small
\begin{quote}
In 3 years, Jayden will be half of Ernesto's age.
If Ernesto is 11 years old, how many years old is Jayden now?
\end{quote}
\normalsize
\paragraph{Answer.}
\small
\begin{quote}
Ernesto = 11 + 3 = \texttt{<<11+3=14>>}14 \\ Jayden = 14/2 = \texttt{<<14/2=7>>}7 in 3 years \\ Now = 7 - 3 = \texttt{<<7-3=4>>}4 \\ Jayden is 4 years old.
\\ \#\#\#\# 4
\end{quote}
\normalsize
\subsection{DAG Formalization for GSM8K}
Unlike ProofWriter, GSM8K does not provide explicit proof graphs.
We therefore construct reasoning DAGs by prompting GPT-5 mini \citep{singh2025openai} to extract structured reasoning steps from each example's \texttt{question} and \texttt{answer}, producing a minimal, faithful DAG whose nodes are necessary reasoning statements and whose edges capture premise-to-conclusion dependencies.

Nodes are defined as concise, grammatically complete statements grounded in the original example.
They may come from the question as necessary facts or constraints, or from the answer as intermediate or final conclusions.
Raw equations and calculator-style expressions are excluded; instead, their results are rewritten as natural-language statements.
Edges encode premise-to-conclusion relations over these nodes, allowing multi-premise dependencies while enforcing acyclicity.
This yields structured reasoning graphs comparable in form to the ProofWriter DAGs while remaining faithful to GSM8K's free-form worked solutions.

The prompt used for DAG formalization is shown in Figure~\ref{fig:gsm8k-dag-prompt}.

\begin{figure*}[t]
\centering
\scriptsize
\begin{promptbox}[title={GSM8K DAG Formalization Prompt Template},left=5pt,right=5pt,top=4pt,bottom=4pt]
\ttfamily
\setlength{\parskip}{1pt}
You are given a math word problem and its worked answer.
Your task is to extract a reasoning DAG from the example.

A reasoning DAG contains:
1. Nodes: meaningful reasoning statements needed to solve the problem.
2. Edges: premise-to-conclusion relations showing how the reasoning progresses.

Your goal is to identify all and only the meaningful reasoning steps relevant to solving the question.

\psection{Important requirements}

1. Node definition

- A node must be a concise, grammatically complete statement in sentence form.
- A node may come from the question if it provides a necessary fact or constraint.
- A node may come from the answer if it states an intermediate conclusion or the final conclusion.
- Include only nodes that are necessary for solving the problem.
- Prefer atomic statements: each node should express exactly one fact or one conclusion whenever possible.
- Every node must be grounded in the input text: it must be either directly stated in the question or answer, or a plain-language restatement of a calculation result already present in the answer.
- Do not create nodes for irrelevant information, external knowledge, unit conversions, common-sense facts, or background facts that are not explicitly stated in the question or answer.

2. Handling calculations

- Do not include raw calculations, equations, arithmetic expressions, or calculator-style text as nodes.
- Rewrite calculation results as natural-language statements.
- For example, instead of ``80/100 * 10 = 8,'' write ``There are 8 more purple flowers than yellow flowers.''

3. Edge definition

- Each edge must contain one or more premise nodes and exactly one conclusion node.
- Represent each edge as:
\{"premises": ["node1", "node2"], "conclusion": "node3"\}
- Create edges only for true reasoning dependencies.
- If a conclusion depends on multiple premises, include all of them.
- The graph must be acyclic.

4. Output rules

- Output exactly one valid JSON object with two keys: ``nodes'' and ``edges''.
- ``nodes'' must be a list of objects of the form:
\{"id": "node1", "text": "..."\}
- ``edges'' must be a list of objects of the form:
\{"premises": ["node1", "node2"], "conclusion": "node3"\}
- Use node IDs in the format node1, node2, node3, ...
- Reason internally as needed, but output only the final JSON object.
- Do not include markdown code fences or any text before or after the JSON.

5. Normalization and quality criteria

- The nodes and edges should be sufficient to reconstruct the reasoning process from premises to final answer.
- Prefer a clean, minimal, and faithful decomposition.
- Preserve the original meaning exactly and do not introduce new assumptions.
- Split combined facts into separate nodes when that makes the reasoning structure clearer.
- Avoid ambiguous pronouns when rewriting node text.
- Include the final answer as a final-conclusion node.
- If a fact from the question is needed later, include it as a node even if it is not repeated in the answer.
- If the answer skips an implicit reasoning step, include it only when it is directly recoverable from information explicitly stated in the question or answer.
- Do not create standalone nodes for unstated background knowledge; such knowledge may help determine edges, but it must not appear as a node.

\psection{Example}

Question:

In 3 years, Jayden will be half of Ernesto's age.
If Ernesto is 11 years old, how many years old is Jayden now?

Answer:

Ernesto = 11 + 3 = \texttt{<<11+3=14>>}14 \\
Jayden = 14/2 = \texttt{<<14/2=7>>}7 in 3 years \\
Now = 7 - 3 = \texttt{<<7-3=4>>}4 \\
Jayden is 4 years old. \\
\#\#\#\# 4

Expected output:

\{"nodes": [
  \{"id": "node1", "text": "Ernesto is 11 years old now."\},
  \{"id": "node2", "text": "In 3 years, Jayden will be half of Ernesto's age."\},
  \{"id": "node3", "text": "In 3 years, Ernesto will be 14 years old."\},
  \{"id": "node4", "text": "In 3 years, Jayden will be 7 years old."\},
  \{"id": "node5", "text": "Jayden is 4 years old now."\}
],
"edges": [
  \{"premises": ["node1"], "conclusion": "node3"\},
  \{"premises": ["node2", "node3"], "conclusion": "node4"\},
  \{"premises": ["node4"], "conclusion": "node5"\}
]\}

Now process the following example.

Question:
\{question\}

Answer:
\{answer\}
\end{promptbox}
\caption{Prompt template used to formalize GSM8K worked solutions as reasoning DAGs.}
\label{fig:gsm8k-dag-prompt}
\end{figure*}

\subsection{Probing Setup and Layerwise Results on GSM8K}
For GSM8K, we restrict the analysis to automatically constructed reasoning DAGs whose sink depth is at least 5, and retain the same probing setup as in ProofWriter.
This filtering aligns graph complexity between GSM8K and ProofWriter by focusing the analysis on examples with sufficiently deep multi-step reasoning structure.

\begin{figure*}[t]
\centering
\includegraphics[width=0.75\textwidth]{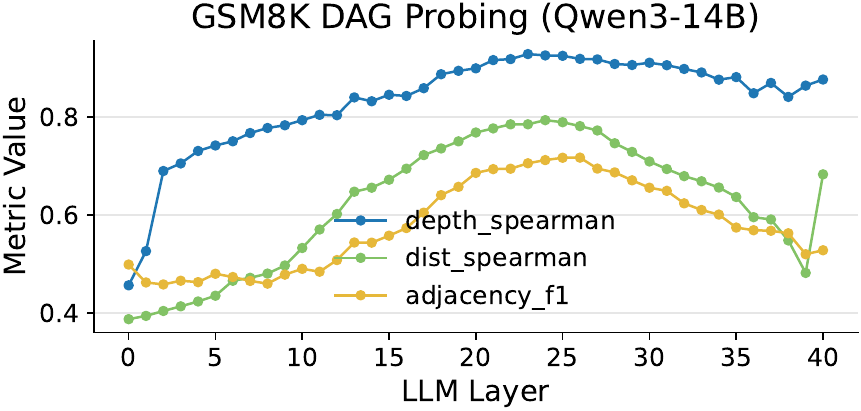}
\caption{Layerwise probing performance on GSM8K for Qwen3-14B.
We report depth Spearman correlation, distance Spearman correlation, and adjacency F1.}
\label{fig:gsm8k_probe}
\end{figure*}

As shown in Figure~\ref{fig:gsm8k_probe}, the layerwise pattern is qualitatively consistent with ProofWriter, with all three metrics rising through early layers, peaking in a broad band of intermediate layers, and declining in later layers.
Absolute probe performance is generally higher than on ProofWriter. One likely contributing factor is that GSM8K statements are more realistic and grounded in real-world regularities, and are less tightly controlled to be mutually hard to distinguish, making reasoning structure easier to encode in hidden states.

\subsection{Baseline Comparison on GSM8K}
\begin{figure*}[t]
\centering
\includegraphics[width=0.75\textwidth]{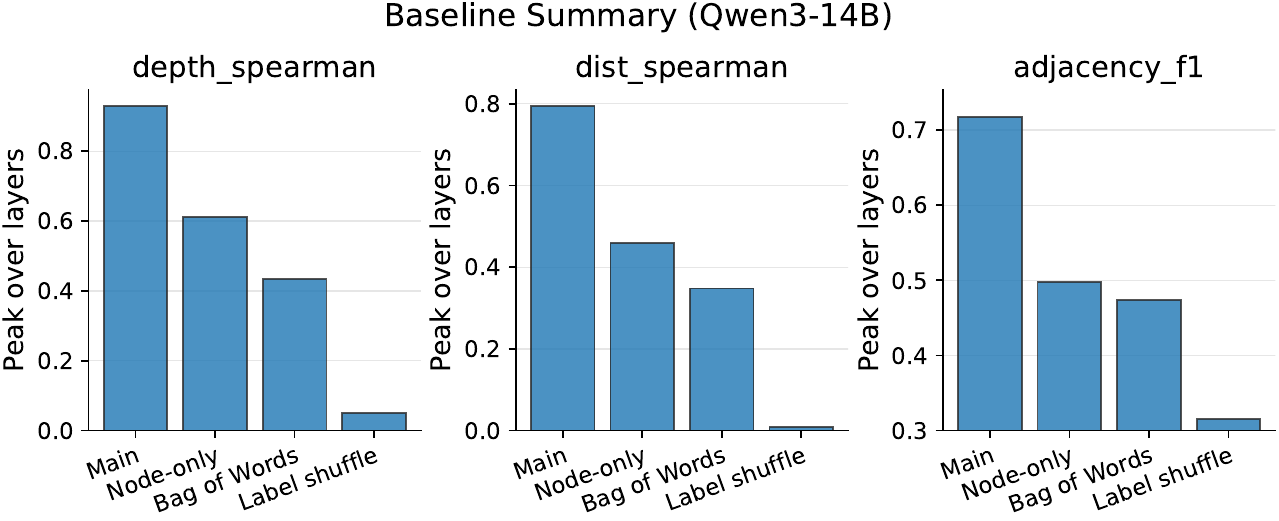}
\caption{Peak probing performance across layers for the main method and baselines on GSM8K using Qwen3-14B.}
\label{fig:gsm8k_baseline}
\end{figure*}

Figure~\ref{fig:gsm8k_baseline} summarizes peak probing performance across layers for each baseline on Qwen3-14B.
As in ProofWriter, the main method remains strongest across depth, distance, and adjacency.
Compared with ProofWriter, the node-only and bag-of-words baselines are slightly stronger, likely because GSM8K statements are less tightly controlled to be mutually hard to distinguish, making surface cues more informative.
At the same time, label shuffling still nearly collapses performance.

\subsection{Layerwise Probing Results Across Models}
\begin{figure*}[t]
\centering
\includegraphics[width=0.48\textwidth]{fig/GSM8K_Qwen3-14B.pdf}
\includegraphics[width=0.48\textwidth]{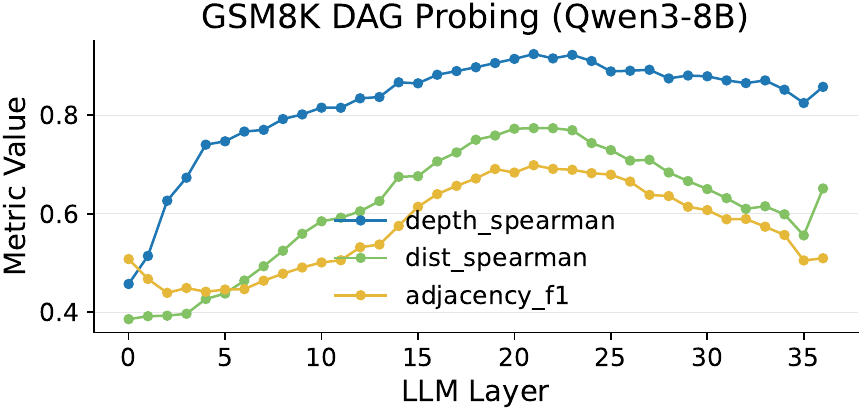}
\includegraphics[width=0.48\textwidth]{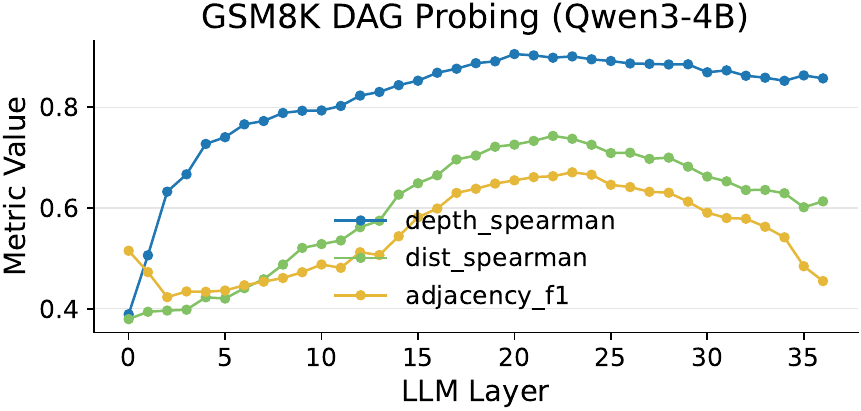}
\includegraphics[width=0.48\textwidth]{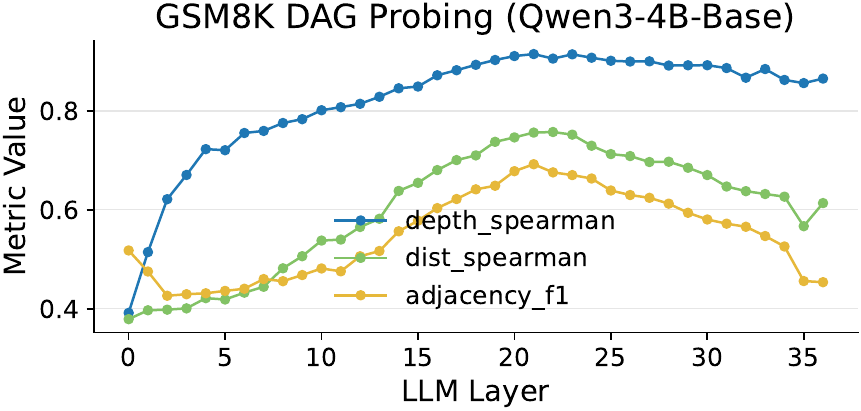}
\includegraphics[width=0.48\textwidth]{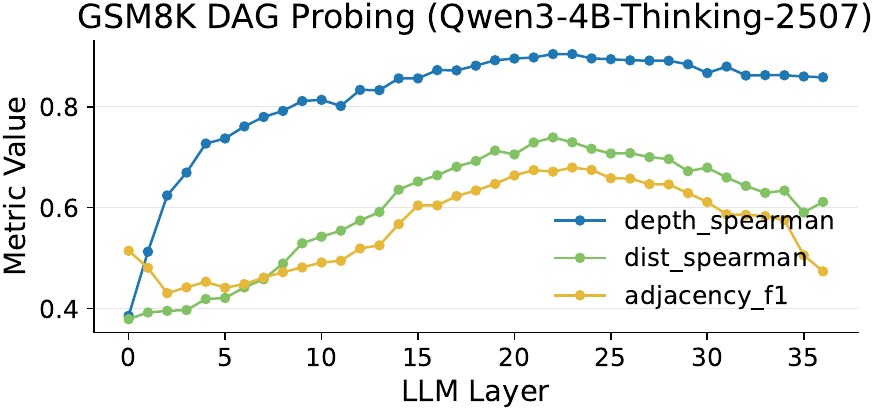}
\includegraphics[width=0.48\textwidth]{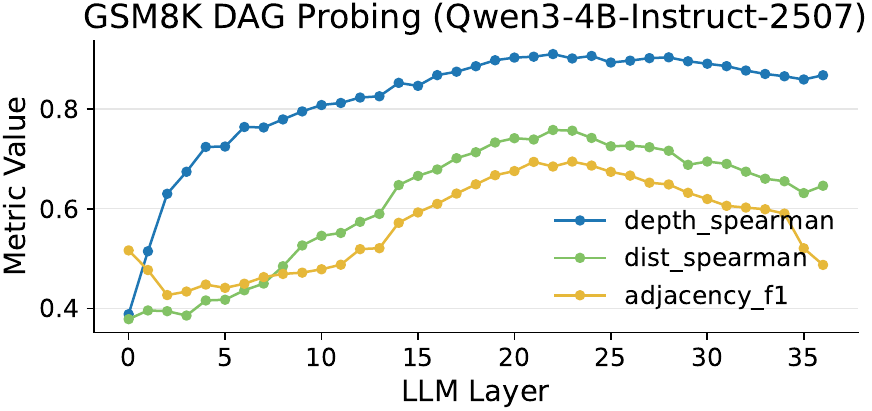}
\includegraphics[width=0.48\textwidth]{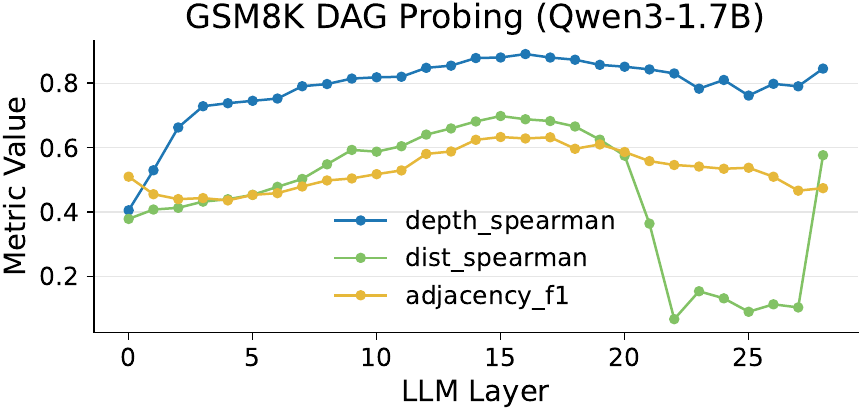}
\includegraphics[width=0.48\textwidth]{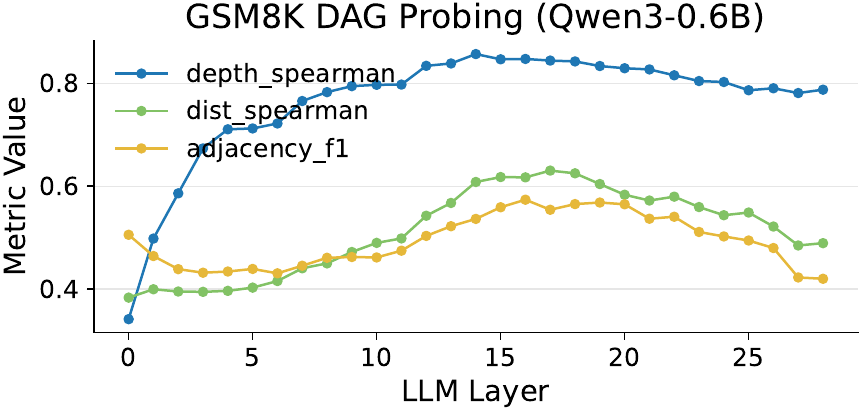}
\caption{Layerwise probing performance on GSM8K across the Qwen3 model family.
Each panel reports depth Spearman correlation, distance Spearman correlation, and adjacency F1 as a function of layer depth. }
\label{fig:layerwise_qwen_gsm8k}
\end{figure*}

\begin{figure*}[t]
\centering
\includegraphics[width=0.48\textwidth]{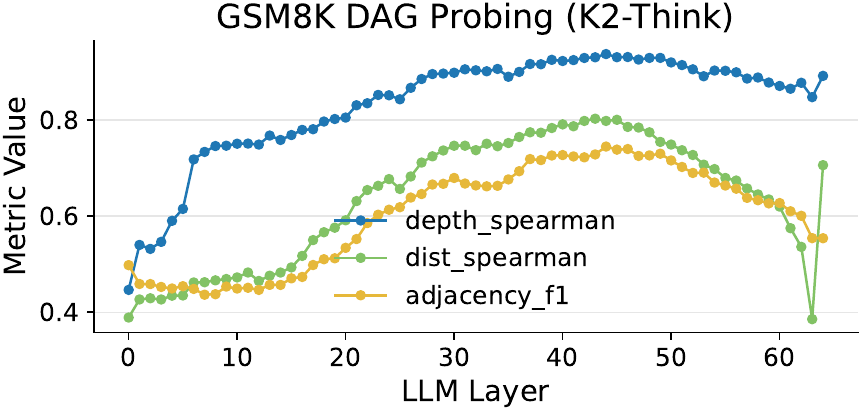}
\includegraphics[width=0.48\textwidth]{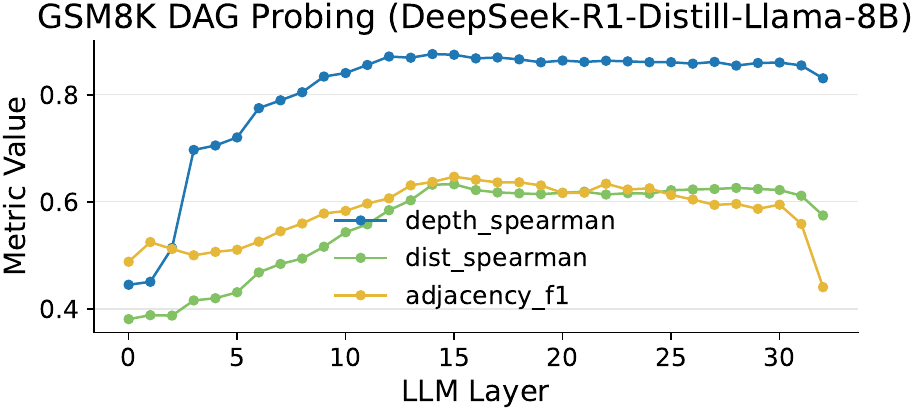}
\includegraphics[width=0.48\textwidth]{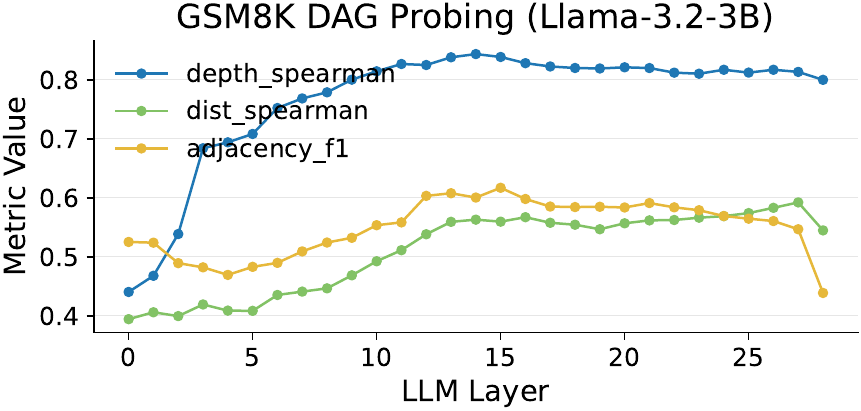}
\caption{Layerwise probing performance on GSM8K across non-Qwen model families.
These plots are included for completeness and show broadly similar qualitative trends across architectures, scales, and training recipes. }
\label{fig:layerwise_nonqwen_gsm8k}
\end{figure*}

We additionally report layerwise probing results for all evaluated models on GSM8K to show that the qualitative emergence pattern is not specific to Qwen3-14B.
Figures~\ref{fig:layerwise_qwen_gsm8k} and \ref{fig:layerwise_nonqwen_gsm8k} generally exhibit the same broad pattern as in ProofWriter and in Qwen3-14B on GSM8K.

\section{Generalization to TACO}
\label{sec:taco}
\subsection{TACO as a Code-generation Testbed}
We use TACO \citep{li2023taco} as an algorithmic code-generation setting for evaluating generation-time DAG recoverability during answer generation.
Unlike ProofWriter and GSM8K, TACO examples are programming problems: each instance provides a natural-language problem statement, reference Python solutions, and input-output tests.
This makes TACO a qualitatively different testbed in which the target reasoning structure is not a logical proof or a worked arithmetic solution, but an algorithmic procedure expressed as executable code.

For generation-time evaluation, we prompt the model with the programming problem and ask it to produce a step-by-step solution followed by a complete Python program.
Correctness is evaluated against the benchmark's input-output tests.
For probing, we associate each selected problem with a reasoning DAG derived from a reference solution, allowing us to ask whether model representations recover algorithmic dependency structure as code is generated.

\subsection{DAG Construction for TACO}
TACO does not provide gold reasoning graphs, so we construct DAGs directly from reference Python solutions.
We use the \texttt{expanded\_code} construction, which treats the program as a structured algorithm rather than a flat sequence of lines.
In the prepared \texttt{expanded\_code} split, we retain EASY and MEDIUM problems whose selected reference solution yields a valid graph with 5--100 nodes and sink depth at least 5, producing 170 training, 30 development, and 60 test examples.
The converter first selects Python-like reference solutions with usable input-output tests, parses each candidate solution with Python's abstract syntax tree (AST), and filters out unsupported or unsafe constructs such as asynchronous control flow, classes, exceptions, unrestricted imports, and recursive self-calls.

Given a valid AST, the converter builds a program-dependence DAG.
Nodes correspond to input reading, assignments, conditionals, loops, returns, expressions, and helper-function computations.
Edges encode data and control dependencies: a statement depends on the most recent producers of the variables it reads, on active control nodes such as surrounding \texttt{if} or loop headers, and on prior statements when no more specific dependency is available.
The final output statement is used as the sink node, and graph depths and pairwise distances are computed from the resulting acyclic dependency graph.

The \texttt{expanded\_code} setting expands two program structures that would otherwise be too coarse.
First, loop headers are not treated as single opaque summaries: the converter recursively processes loop bodies and attaches their statements to the loop-control node.
Second, calls to helper functions are inlined at the graph level.
For each helper call, the converter creates a helper-entry node, maps call arguments to function parameters, recursively processes the helper body under the call context, and then records the helper result as an additional premise for the calling statement.
A call stack prevents recursive expansion, preserving acyclicity.
This produces DAGs whose nodes expose both local program steps and longer-range dependencies through loop bodies and helper calls, making them suitable for the same probing pipeline used for ProofWriter and GSM8K.

\section{Compute Infrastructure}
All experiments use frozen pretrained models; only lightweight linear probes are trained on cached hidden representations.
The evaluated model sizes are reported by the model names in the main text and appendix.
The full pipeline requires at least two NVIDIA A40 GPUs for the largest evaluated models, and we used up to eight A40 GPUs concurrently to parallelize representation extraction and accelerate iteration.

\end{document}